\documentclass[nohyperref]{article}

\usepackage{microtype}
\usepackage{graphicx}
\usepackage{graphbox}
\usepackage{subfigure}
\usepackage{booktabs} 
\usepackage{multirow}
\usepackage{multicol}
\usepackage{textcomp} 
\usepackage[inline]{enumitem}
\usepackage{algorithm}
\usepackage{algpseudocode}
\usepackage{amsthm}

\usepackage{hyperref}
\usepackage{relsize}
\usepackage{array}

\usepackage[accepted]{icml2022}

\usepackage{amsmath}
\usepackage{amssymb}
\usepackage{mathtools}
\usepackage{amsthm}

\DeclareMathOperator*{\argmin}{arg\,min}

\definecolor{TartOrange}{HTML}{ff2e35}
\definecolor{Orange}{HTML}{ff7825}
\definecolor{Mango}{HTML}{ffc013}
\definecolor{AppleGreen}{HTML}{7cb81b}
\definecolor{Blue}{HTML}{1173b0}
\definecolor{BdazzledBlue}{HTML}{2e58a5}
\definecolor{Purple}{HTML}{5b3590}
\definecolor{Sunglow}{HTML}{FFCA3A}

\newcommand{\addcomment}[3]{} 

\renewcommand{\thesubfigure}{(\Alph{subfigure})}
\newcommand{\tabcaption}{\refstepcounter{subfigure}\thesubfigure}

\usepackage[capitalize,noabbrev]{cleveref}

\theoremstyle{plain}

\theoremstyle{definition}

\theoremstyle{remark}

\newcolumntype{L}[1]{>{\centering\arraybackslash}m{#1}}
\newcolumntype{C}[1]{>{\raggedright\arraybackslash}m{#1}}

\usepackage[textsize=tiny]{todonotes}

\icmltitlerunning{Hindering Adversarial Attacks with Implicit Network Activation Coding (LINAC)}

\usepackage{xspace}
\newcommand{\privatekey}{\emph{private~key}\xspace}
\newcommand{\autoattack}{AA}
\newcommand{\mt}{MT}
\newcommand{\pgd}{PGD}
\newcommand{\sqr}{Square}
\newcommand{\bestknown}{\textit{Best Known}\xspace}
\newcommand{\bestadversary}{\textit{Best Adversary}\xspace}

\newcommand{\mlpparam}{\phi}
\newcommand{\mlp}{\Phi}
\newcommand{\pos}{\mathbf{p}}
\newcommand{\from}{\colon}

\begin{document}

\twocolumn[
\icmltitle{Hindering Adversarial Attacks with Implicit Neural Representations}



\icmlsetsymbol{equal}{*}

\begin{icmlauthorlist}
\icmlauthor{Andrei A. Rusu}{comp}
\icmlauthor{Dan A. Calian}{comp}
\icmlauthor{Sven Gowal}{comp}
\icmlauthor{Raia Hadsell}{comp}
\end{icmlauthorlist}

\icmlaffiliation{comp}{DeepMind, London, UK}

\icmlcorrespondingauthor{Andrei A. Rusu}{andrei@deepmind.com}

\icmlkeywords{Machine Learning, ICML}

\vskip 0.3in
]



\printAffiliationsAndNotice{}  

\begin{abstract}
We introduce the Lossy Implicit Network Activation Coding (LINAC) defence, an input transformation which successfully hinders several common adversarial attacks on CIFAR-$10$ classifiers for perturbations up to $\epsilon = 8/255$ in $L_\infty$ norm and $\epsilon = 0.5$ in $L_2$ norm. Implicit neural representations are used to approximately encode pixel colour intensities in $2\text{D}$ images such that classifiers trained on transformed data appear to have robustness to small perturbations without adversarial training or large drops in performance. The seed of the random number generator used to initialise and train the implicit neural representation turns out to be necessary information for stronger generic attacks, suggesting its role as a \privatekey. We devise a Parametric Bypass Approximation (PBA) attack strategy for key-based defences, which successfully invalidates an existing method in this category. Interestingly, our LINAC defence also hinders some transfer and adaptive attacks, including our novel PBA strategy. Our results emphasise the importance of a broad range of customised attacks despite apparent robustness according to standard evaluations. \textit{LINAC source code and parameters of defended classifier evaluated throughout this submission are available: \href{https://github.com/deepmind/linac}{https://github.com/deepmind/linac}.}
\end{abstract}

\section{Introduction}

Training Deep Neural Network (DNN) classifiers which are accurate yet generally robust to small adversarial perturbations is an open problem in computer vision and beyond, inspiring much empirical and foundational research into modern DNNs. \citet{Szegedy2014intriguing} showed that DNNs are not inherently robust to imperceptible input perturbations, which reliably cross learned decision boundaries, even those of different models trained on similar data. With hindsight, it becomes evident that two related yet distinct design principles have been at the core of proposed defences ever since. Intuitively, accurate DNN classifiers could be considered robust in practice if:~\begin{enumerate*}[label=(\Roman*)]
\item their decision boundaries were largely insensitive to \emph{all} adversarial perturbations, and/or\label{enumerate:insensitive}
\item computing \emph{any} successful adversarial perturbations was shown to be expensive, ideally intractable\label{enumerate:intractable} 
\end{enumerate*}. Early defences built on principle \ref{enumerate:insensitive} include the adversarial training approach of \citet{Madry2018at} and the verifiable defences of \citet{Hein2017formal, Raghunathan2018certified}, with many recent works continually refining such algorithms, e.g. \citet{Cohen2019certified, Gowal2020uncovering, Rebuffi2021fixing}. A wide range of defences were built, or shown to operate, largely on principle \ref{enumerate:intractable}, including adversarial detection methods \citep{Carlini2017adversarial}, input transformations \citep{Guo2018countering} and denoising strategies \citep{Liao2018defense, Niu2020limitations}. Many such approaches have since been circumvented by more effective attacks, such as those proposed by \citet{CarliniWagner2017towards}, or by using ``adaptive attacks'' \citep{Athalye2018obfuscated, Tramer2020adaptive}.

Despite the effectiveness of recent attacks against these defences,  \citet{Garg2020adversarially} convincingly argue on a theoretical basis that principle \ref{enumerate:intractable} is sound; similarly to cryptography, robust learning could rely on computational hardness, even in cases where small adversarial perturbations do exist and would be found by a hypothetical, computationally unbounded adversary. However, constructing such robust classifiers for problems of interest, e.g. image classification, remains an open problem. Recent works have proposed defences based on cryptographic principles, such as the pseudo-random block pixel shuffling approach of \citet{Aprilpyone2021block}. As we will show, employing cryptographic principles in algorithm design  is not in itself enough to prevent efficient attacks. Nevertheless, we build on the concept of key-based input transformation and propose a novel defence based on Implicit Neural Representations (INRs). We demonstrate that our Lossy Implicit Neural Activation Coding (LINAC) defence hinders most standard and even adaptive attacks, more so than the related approaches we have tested, without making any claims of robustness about our defended classifier.

{\textbf{Contributions:}}
\begin{enumerate*}[label=(\arabic*)]
\item We demonstrate empirically that lossy INRs can be used in a standard CIFAR-10 image classification pipeline if they are computed using \emph{the same implicit network initialisation}, a novel observation which makes our LINAC defence possible.
\item The seed of the random number generator used for initialising and computing INRs is shown to be an effective and compact \privatekey, since withholding this information hinders a suite of standard adversarial attacks widely used for robustness evaluations.
\item We report our systematic efforts to circumvent the LINAC defence with transfer and a series of adaptive attacks, designed to expose and exploit potential weaknesses of LINAC.
\item To the same end we propose the novel Parametric Bypass Approximation (PBA) attack strategy, valid under our threat model, and applicable to other defences using secret keys. We demonstrate its effectiveness by invalidating an existing key-based defence which was previously assumed robust.
\end{enumerate*}

\begin{figure*}[]
\centering
\includegraphics[width=0.85\linewidth]{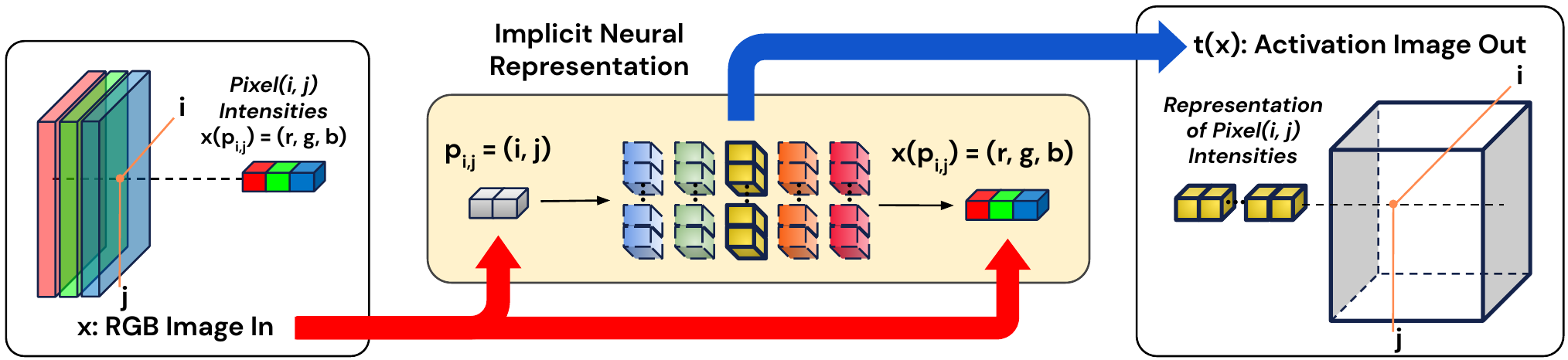}

\caption{Visual depiction of LINAC, our proposed input transformation. An RGB image $x$ is converted into an \emph{Activation Image} $t(x)$ with identical spatial dimensions, but $H$ channels instead of $3$. A neural network model which maps pixel coordinates to RGB colour intensities is fit such that it approximates $x$. The resulting model parameters (after fitting) are called the Implicit Neural Representation (INR) of  image $x$. In order to output correct RGB colour intensities for all pixels, the implicit neural network needs to compute a hierarchical functional decomposition of $x$. We empirically choose an intermediate representation to define our transformation. \emph{Activations} in the middle hidden layer are associated with their corresponding pixel coordinates to form the output \emph{Activation Image} $t(x)$, with as many channels as there are units in the middle layer (H).\label{figs:representation}}
\end{figure*}

\section{Related Work}

{\textbf{Adversarial Robustness.}} Much progress has been made towards robust image classifiers along the adversarial training \citep{Madry2018at} route, which has been extensively explored and is well reviewed, e.g. in \cite{Schott2019towards, Pang2020bag, Gowal2020uncovering, Rebuffi2021fixing}. While such approaches can be effective against current attacks, a complementary line of work investigates certified defences, which offer guarantees of robustness around examples for some well defined sets \citep{Wong2018provable, Raghunathan2018certified, Cohen2019certified}. Indeed, many such works acknowledge the need for complementary approaches, irrespective of the success of adversarial training and the well understood difficulties in combining methods  \citep{He2017adversarial}. The prolific work on defences against adversarial perturbations has spurred the development of stronger attacks \citep{CarliniWagner2017towards, Brendel2018decision, Andriushchenko2020square} and standardisation of evaluation strategies for  threat models of interest \citep{Athalye2018obfuscated, CroceHein2020reliable}, including adaptive attacks \citep{Tramer2020adaptive}. Alongside the empirical progress towards building robust predictors, this line of research has yielded an improved understanding of current deep learning models \citep{Ilyas2019bugs, Engstrom2019adversarial}, the limitations of effective adversarial robustness techniques \citep{Jacobsen2018excessive}, and the data required to train them \citep{Schmidt2018adversarially}.

\citet{Athalye2018obfuscated} show that a number of defences primarily hinder gradient-based adversarial attacks by \emph{obfuscating gradients}. Various forms are identified, such as gradient shattering \citep{Goodfellow2014explaining}, gradient masking \citep{Papernot2017practical}, exploding and vanishing gradients \citep{Song2018pixeldefend}, stochastic gradients \citep{Dhillon2018stochastic} and a number of input transformations aimed at countering adversarial examples, including noise filtering approaches using PCA or image quilting \citep{Guo2018countering}, the Saak transform \citep{Song2018defense}, low-pass filtering \citep{Shaham2018defending}, matrix estimation \citep{Yang2019me} and JPEG compression \citep{Dziugaite2016study, Das2017keeping, Das2018shield}. Indeed, many such defences have been proposed, as reviewed by \citet{Niu2020limitations}, they have ranked highly in competitions \citep{Kurakin2018adversarial}, and many have since been shown to be less robust than previously thought, e.g. by \citet{Athalye2018obfuscated} and  \citet{Tramer2020adaptive}, who use adaptive attacks to demonstrate that several input transformations offer little to no robustness.

To build on such insights, it is worth identifying the ``ingredients'' essential to the success of adversarial attacks. Most effective attacks, including adaptive ones, assume the ability to approximate the outputs of the targeted model for arbitrary inputs. This is reasonable when applying the correct transformation is tractable for the attacker. Hence, \emph{denying access} to such computations seems to be a promising direction for hindering adversarial attacks. \citet{Aprilpyone2020extension, Aprilpyone2021transfer, Maungmaung2021protection} borrow standard practice from cryptography and assume that an attacker has full knowledge of the defence's algorithm and parameters, short of a small number of bits which make up a \privatekey. Another critical ``weakness'' of such input denoising defences is that they can be approximated by the identity mapping for the purpose of computing gradients \citep{Athalye2018obfuscated}. Even complex parametric approaches, which learn stochastic generative models of the input distribution, are susceptible to reparameterisation and Expectation-over-Transformation (EoT) attacks in the white-box setting. Thus, it is worth investigating whether \emph{non-parametric}, \emph{lossy} and \emph{fully deterministic} input transformations exist such that downstream models can still perform tasks of interest to high accuracy, while known and novel attack strategies are either ruled out, or at least substantially hindered, including adaptive attacks.

{\textbf{Implicit Neural Representations.}}
Neural networks have been used to parameterise many kinds of signals, see the work by \citet{Sitzmann2020list} for an extensive list, with remarkable recent advances in scene representations \citep{Mildenhall2020nerf} and image processing \citep{Sitzmann2020implicit}. INRs have been used in isolation per image or scene, not for generalisation across images. Some exceptions exist in unsupervised learning, e.g. \citet{Skorokhodov2021adversarial} parameterise GAN decoders such that they directly output INRs of images,  rather than colour intensities for all pixels. In this paper we show that INRs can be used to discover functional decompositions of RGB images which enable comparable generalisation to learning on the original signal encoding (i.e.\ RGB).

\section{Hindering Adversarial Attacks with Implicit Neural Representations}

In this section we introduce LINAC, our proposed input transformation which hinders adversarial attacks by leveraging implicit neural representations, also illustrated in Fig.~\ref{figs:representation}.

{\bf Setup.} We consider a supervised learning task with a dataset $\mathcal{D} \subset \mathcal{X} \times \mathcal{Y}$ of pairs of images $x$ and their corresponding labels $y$. We use a deterministic input transformation $t \from \mathcal{X} \to \mathcal{H}$ which transforms input images, $x \mapsto t(x)$, while preserving their spatial dimensions. Further, we consider a classifier $f_\theta$, parameterised by $\theta$, whose parameters are estimated by Empirical Risk Minimisation (ERM) to map transformed inputs to labels $f_\theta \from \mathcal{H} \to \mathcal{Y}$. The model is not adversarially trained, yet finding adversarial examples for it is hindered by LINAC, as we demonstrate through extensive evaluations in Section \ref{sec:results}.

{\bf Implicit Neural Representations.} For an image $x$, its implicit neural representation is given by a multi-layer perceptron (MLP) $\mlp = h^{L} \circ h^{L-1} \circ \cdots \circ h^0$, $\mlp \from \mathbb{R}^2 \to \mathbb{R}^3$, with $L$ hidden layers, which maps spatial coordinates to their corresponding colours. $\mlp_\mlpparam$ is a solution to the implicit equation:
\begin{equation}\label{eq:implicit}
\mlp(\pos) - x(\pos) = 0,
\end{equation}
where $\pos$ are spatial coordinates (i.e.\ pixel locations) and $x(\pos)$ are the corresponding image colours. Our input transformation leverages this implicit neural representation to encode images in an approximate manner.


\begin{algorithm}[tb]
\caption{The LINAC Transform\label{alg:linac}}
\begin{algorithmic}
\State \textbf{Inputs:} RGB image $x$ (with size $I \times J \times 3$); \privatekey; number of epochs $N$; mini-batch size $M$; number of MLP layers $L$; representation layer $K$; learning rate $\mu$.
\State \textbf{Output:} Activation Image $t(x)$ (with size $I \times J \times H$).

\State $\text{rng} = \textsc{init\_prng}(\privatekey)$ \Comment{Seed rng.}
\State $\mlpparam^{(0)} = \textsc{init\_mlp}(\text{rng}, L)$ 
\State $S = \lfloor I \cdot J / M\rfloor$  \Comment{Num. mini-batches per epoch.}

\State $\mlpparam = \mlpparam^{(0)}$
\For{$epoch=0 \ldots N-1$}
    \State $\mathcal{P} = \textsc{shuffle\_and\_split\_pixels}(x, \text{rng}, S)$ 
    \For{$m=0 \ldots S-1$}
        \State $\ell = \frac{1}{M \cdot I \cdot J}\sum\limits_{(i,j) \in \mathcal{P}[m]} || \mlp_\mlpparam(\pos_{i, j}) - x(\pos_{i, j}) ||_2^2$
        \State $\mlpparam = \mlpparam - \mu \nabla_\mlpparam \ell$
    \EndFor
\EndFor
\State $\hat{\mlpparam}_{x} = \mlpparam$
\State {\bf Return} $t(x)$ applying Eq.~\ref{eq:transformation} using $\hat{\mlpparam}_{x}$ and layer $K$.

\end{algorithmic}
\end{algorithm}


{\bf Reconstruction Loss.} The implicit equation (\ref{eq:implicit}) can be translated~\citep{Sitzmann2020implicit} into a standard reconstruction loss between image colours and the output of a multi-layer perceptron $\mlp_\mlpparam$ at each (2D) pixel location $\pos_{i, j}$,
\begin{equation}
\label{eq:loss}
\mathcal{L}(\mlpparam, x) = \sum_{i, j} || \mlp_\mlpparam(\pos_{i,j}) - x(\pos_{i, j}) ||_2^2.
\end{equation}

We provide pseudocode for the LINAC transform in Algorithm~\ref{alg:linac} and a discussion of computational and memory requirements in Appendix~\ref{app:compreq}. For each individual image $x$, we estimate $\hat{\mlpparam}_x$, an approximate local minimiser of $\mathcal{L}(\mlpparam, x)$, using a stochastic iterative minimisation procedure with mini-batches of $M$ pixels grouped into epochs, which cover the entire image in random order, for a total of $N$ passes through all pixels. 

{\bf Private Key.} A random number generator is used for: (1) generating the initial MLP parameters $\mlpparam^{(0)}$ and (2) for deciding which random subsets of pixels make up mini-batches in each epoch. This random number generator is seeded by a $64$-bit integer which we keep secret and denote as the \privatekey. Hence, for all inputs $x$ we start each independent optimisation from the same set of initial parameters $\mlpparam^{(0)}$, and we use the same shuffling of pixels across epochs.

\newpage

{\bf Lossy Implicit Network Activation Coding (LINAC).} We consider the lossy encoding of each pixel $(i, j)$ in image $x$ as the $H$-dimensional intermediate activations vector of layer $K$ of the MLP evaluated at that pixel position: $c_x(i, j) = (h^{K-1}_{\hat{\mlpparam}_{x}} \circ \cdots \circ h^0_{\hat{\mlpparam}_{x}})(\pos_{i,j})$ with $K < L$. We build the \textit{lossy implicit network activation coding} transformation of an image $x$ by stacking together the encodings of all its pixels in its 2D image grid, concatenating on the feature dimension axis. The LINAC transformation $t(x)$ of the $I \times J \times 3$ image $x$ is given by: \begin{equation}\label{eq:transformation}
t(x) = \begin{bmatrix} 
    c_x({0,0})      & \dots     & c_x({0,J-1})  \\
    \vdots          & \ddots    & \vdots        \\
    c_x({I-1,0})    & \dots     & c_x({I-1,J-1})
    \end{bmatrix},
\end{equation}
and has dimensionality $I \times J \times H$, where $H$ is the number of outputs of the $K$-th layer of the MLP.
By construction, our input transformation preserves the spatial dimensions of each image while increasing the feature dimensionality (from $3$, the image's original number of colour channels, to $H$); this means that standard network architectures used for image classification (e.g. convolutional models) can be readily trained as the classifier $f_\theta$.

All omitted implementation details are provided in Appendix~\ref{sec:impl_details}, and sensitivity analyses of LINAC to its hyper-parameters are reported in Appendix~\ref{sec:sensitivity_analysis}.

{\bf Threat Model.} We are interested in hindering adversarial attacks on a nominally-trained classifier $f_\theta(t(x))$, which operates on transformed inputs (i.e.\ on $t(x)$ rather than on $x$), using a \privatekey of our choosing. Next, we describe the threat model of interest by stating the conditions under which the LINAC defence is meant to hinder adversarial attacks on $f_\theta$, following \citet{Aprilpyone2021block}.

We assume attackers do not have access to the \privatekey, the integer seed of the random number generator used for computing the LINAC transformation, but otherwise have full algorithmic knowledge about our approach. Specifically, we assume an attacker has complete information about the classification pipeline, including the architecture, training dataset and weights of the defended classifier. This includes full knowledge of the LINAC algorithm, the implicit network architecture, parameter initialisation scheme and all the fitting details, except for the \privatekey.


\section{Attacking the LINAC Defence}

{\bf Setup.} We are interested in evaluating the apparent robustness of a LINAC-defended classifier, $f_{\hat{\theta}}$, which has been trained by ERM to classify transformed inputs from the dataset $\mathcal{D}$. Specifically, its parameters $\hat{\theta}$ minimise $\mathbb{E}_{x, y \sim \mathcal{D}}\left[  \mathcal{L}_\text{CE}(f_\theta(t(x)), y) \right]$ , where $\mathcal{L}_\text{CE}$ is the cross-entropy loss and $t(x)$ is the LINAC transformation applied to image $x$ using the \privatekey.

{\bf Input Perturbations.} Classifiers defended by LINAC are not adversarially trained~\citep{Madry2018at} to increase their robustness to specific $L_p$ norm-bounded input perturbations. 
Furthermore, the LINAC defence is inherently agnostic about particular notions of maximum input perturbations. Nevertheless, to provide results comparable with a broad set of defences from the literature, we perform evaluations on standard $L_p$ norm-bounded input perturbations with: (1) a maximum perturbation radius of $\epsilon = 8 / 255$ in the $L_\infty$ norm, and (2) one of $\epsilon = 0.5$ in the $L_2$ norm.

{\bf Adapting Existing Attacks.} Without access to the \privatekey an attacker cannot compute the LINAC transformation exactly. However, an attacker could acquire access to model inferences by attempting to brute-force guess the \privatekey. Another option would be to train surrogate models with LINAC, but using keys chosen by the attacker, in the hope that decision boundaries of these models would be similar enough to mount effective transfer attacks. More advanced attackers could modify LINAC itself to enable strong Backward Pass Differentiable Approximation (BPDA)~\citep{Athalye2018obfuscated} attacks. We evaluate the success of these and other standard attacks in Section~\ref{sec:results}.

{\bf Designing Adaptive Attacks.} \citet{Athalye2018obfuscated} provide an excellent set of guidelines for designing and performing successful adaptive attacks, while also standardising results reporting and aggregation. Of particular interest for defences based on input transformations are the BPDA and Expectation-over-Transformation (EoT) attack strategies. Subsequent work convincingly argues that adaptive attacks are not meant to be general, and must be customised, or ``adapted'', to each defence in turn \citep{Tramer2020adaptive}. While BPDA and EoT generate strong attacks on input transformations, they both rely on being able to compute the forward transformation or approximate it with samples. Indeed, the authors mention that substitution of both the forward and backward passes with approximations leads to either completely ineffective, or much less effective attacks.

{\bf Parametric Bypass Approximation (PBA)\label{subsec:PBA}.}
Inspired by the reparameterisation strategies of \citet{Athalye2018obfuscated}, we propose a bespoke attack by making use of several pieces of information available under our threat model: the parametric form of the defended classifier $f_{\theta}(t(x))$, its training dataset $\mathcal{D}$ and loss function $\mathcal{L}_{\text{CE}}$, and its trained weights $\hat{\theta}$.

A Parametric Bypass Approximation of an unknown nuisance transformation $u \from \mathcal{X} \to \mathcal{H}$ is a surrogate parametric function $h_{\psi} \from \mathcal{X} \to \mathcal{H}$, parameterised by a solution to the following optimisation problem:
\begin{equation}
\psi^{*} = \underset{\psi}{\argmin}\mathop{\mathbb{E}}_{x, y \sim \mathcal{D}} \left[ \quad \mathcal{L}_\text{CE}(f_{\hat{\theta}}(h_{\psi}(x)), y) \quad \right]. \end{equation}

This formulation seeks a set of parameters $\psi^{*}$ which minimise the original classification loss while keeping  the defended classifier's parameters frozen at $\hat{\theta}$. Similar with classifier training, this optimisation problem can be solved efficiently using Stochastic Gradient Descent (SGD).

A PBA adversarial attack can then proceed by approximating the \emph{defended classifier}  $f_{\hat{\theta}}(u(\cdot))$ with those of the \emph{bypass classifier} $f_{\hat{\theta}}(h_{\psi^{*}}(\cdot))$ in both forward and backward passes when computing adversarial examples, e.g. using Projected Gradient Descent (PGD).

The main advantages of the PBA strategy are that no forward passes through the nuisance transformation $u(\cdot)$ are required, and that it admits efficient computation of many attacks to $f_{\hat{\theta}}$, including gradient-based ones. In Section~\ref{sec:results} we demonstrate the effectiveness of PBA beyond the LINAC defence. We show that, even though the surrogate transformation is fit on training data only, the defended classifier operating on samples passed through $h_{\psi^{*}}$ (bypassing $u$) demonstrates nearly identical generalisation to the test set. Furthermore, we also show that PBA has greater success at finding adversarial examples for the LINAC defence compared to other methods. Lastly, we use PBA to invalidate an existing key-based defence proposed in the literature.

\section{Results}\label{sec:results}
\subsection{Evaluation Methodology}
Since LINAC makes no assumptions about adversarial perturbations, we are able to evaluate \emph{a single defended classifier model} against all attack strategies considered, in contrast to much adversarial training research \citep{Madry2018at}.

To obtain a more comprehensive picture of apparent robustness we start from the rigorous evaluation methodology used by \citet{Gowal2019alternative, Rebuffi2021fixing}. We perform untargeted \pgd\ attacks with $100$ steps and $10$ randomised restarts, as well as multi-targeted (\mt) PGD attacks using $200$ steps and $20$ restarts. Anticipating the danger of obfuscated gradients skewing results, we also evaluate with the Square approach of \citet{Andriushchenko2020square}, a powerful gradient-free attack, with $10000$ evaluations and $10$ restarts. For precise comparisons with the broader literature we also report evaluations using the parameter-free AutoAttack (\autoattack) strategy of \citet{CroceHein2020reliable}.

Following \citet{Athalye2018obfuscated} we aggregate results across attacks by only counting as accurate robust predictions those test images for which the defended classifier predicts the correct class with and without adversarial perturbations, computed using all methods above. We report this as \bestknown robust accuracy.

In instances where several surrogate models are used to compute adversarial perturbations, also known as \emph{transfer attacks}, we report \bestadversary results aggregated for each individual attack, which is defined as robust accuracy against all source models considered.

We aggregate evaluations across these two dimensions (attacks \& surrogate models) by providing a single robust accuracy number against all attacks computed using all source models for each standard convention of maximum perturbation and norm, enabling easy comparisons with results in the literature.

\begin{table*}[t]
\centering%
\resizebox{.8\textwidth}{!}{%
\begin{tabular}{|L{0.6cm}|L{1.75cm}|L{1.2cm}L{1.5cm}L{1.5cm}L{2.5cm}L{2.7cm}|L{2.2cm}|}
\hline
\multicolumn{2}{|c|}{}  & \multicolumn{5}{c|}{\bf Transfer Attack Source Models}  & {\bestadversary}
\\ \hline
{\scriptsize{\bf Norm}} & {\bf Attack}  & {\bf Nominal Source} & {\bf Adversarial Training $(\mathbf{L_\infty})$}  & {\bf Adversarial Training $(\mathbf{L_2})$} & {\bf Defended Surrogates (Attacker Keys)} & {\bf Reconstruction-Based Surrogates (BPDA)} &   {\bf All Source Models} \\
\hline
\hline
\multirow{6}{*}{$L_\infty$}
        &	\autoattack	        &	92.77	&   80.42	&	70.29	&   84.00   &   59.40  &   41.18   \\
        &	\mt     	        &	84.57	&   72.96	&	56.08	&   85.70   &   55.37  &   47.91   \\
        &	\pgd    	        &	85.99	&   60.97	&	44.06	&   87.32   &   56.00  &   41.22   \\
        &	\sqr	            &	85.12	&   65.69	&	52.66	&   75.91   &   69.14  &   49.76   \\
        \cline{2-8}
        &   \bestknown          &   81.91   &   54.97   &   39.20   &   75.64   &   51.17  &   37.04   \\
\hline
\hline
\multirow{6}{*}{$L_2$}      
        &	\autoattack	        &   90.84	&	86.75	&	80.83	&   88.27   &   74.59  &	71.32   \\
        &	\mt     	        &   87.55	&	85.34	&	84.81	&   87.31   &   74.98  &	73.83   \\
        &	\pgd    	        &   88.61	&	82.39	&	74.19	&   88.36   &   75.00  &	70.90   \\
        &	\sqr	            &   88.58	&	84.50	&	79.31	&   84.08   &   83.26  &	77.68   \\
        \cline{2-8}
        &   \bestknown          &   86.06   &   79.42   &   71.92   &   83.48   &   71.89  &    68.41   \\
        \hline
\end{tabular}%
}
\caption{CIFAR-10 test set robust accuracy ($\%$) of a single LINAC defended classifier according to a suite of $L_\infty$ and $L_2$ transfer attacks, valid under our threat model, using various source classifiers to generate adversarial perturbations.\label{table:black-box}}
\end{table*}

\subsection{Attacks with Surrogate Transformations \& Models}
A majority of adversarial attack strategies critically depend on approximating the outputs of the defended classifier for inputs chosen by the attacker. The \privatekey is kept secret in our threat model, which means that an attacker can neither compute the precise input transformation used to train the defended classifier, nor its outputs on novel data. Hence, an attacker must find appropriate surrogate transformations, or surrogate classifier models, in order to perform effective adversarial attacks. We investigate both strategies below.

\begin{figure}[t]
\begin{center}
\includegraphics[width=0.85\linewidth, trim={35px 35px 30px 10px}]{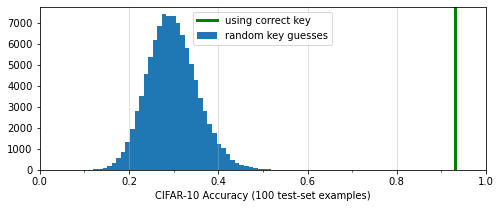}
\end{center}
\caption{Results of direct attack on \privatekey. A histogram of accuracies of the same defended classifier with inputs transformed using either the correct key or $100000$ randomly chosen keys. An appropriate surrogate transformation is not found, invalidating attack vectors which rely on access to the outputs of the defended model on attacker chosen inputs.\label{figs:key_attack}}
\end{figure}

Firstly, we empirically check that the outputs of the defended classifier cannot be usefully approximated without knowledge of the \privatekey. It is reasonable to hypothesise that transformations with different keys may lead to similar functional representations of the input signal. We start investigating this hypothesis by simply computing the accuracy of the defended model on clean input data transformed with LINAC, but using keys chosen by the attacker, also known as a brute-force key attack, which is valid under our threat model. As reported in Figure~\ref{figs:key_attack}, the accuracy of our LINAC defended classifier on test inputs transformed with the correct \privatekey is over $93\%$. In an attempt to find a surrogate transformation, $100000$ keys are picked uniformly at random. For each key, we independently evaluated the accuracy of the classifier using a batch of $100$ test examples, and we report the resulting accuracy estimates for all keys with a histogram plot. The mean accuracy with random key guesses is around $30\%$, with a top accuracy of just $57\%$ (see Table~\ref{table:best_keys} in Appendix~\ref{sec:surrogates} for a breakdown). Hence, using LINAC with incorrect keys leads to poor approximations of classifier outputs on correctly transformed data. This suggest that the learned decision boundaries of the defended classifier are not invariant to the \privatekey used by LINAC.

\begin{figure}[h]
\begin{center}
\includegraphics[width=0.85\linewidth, trim={35px 35px 30px 10px}]{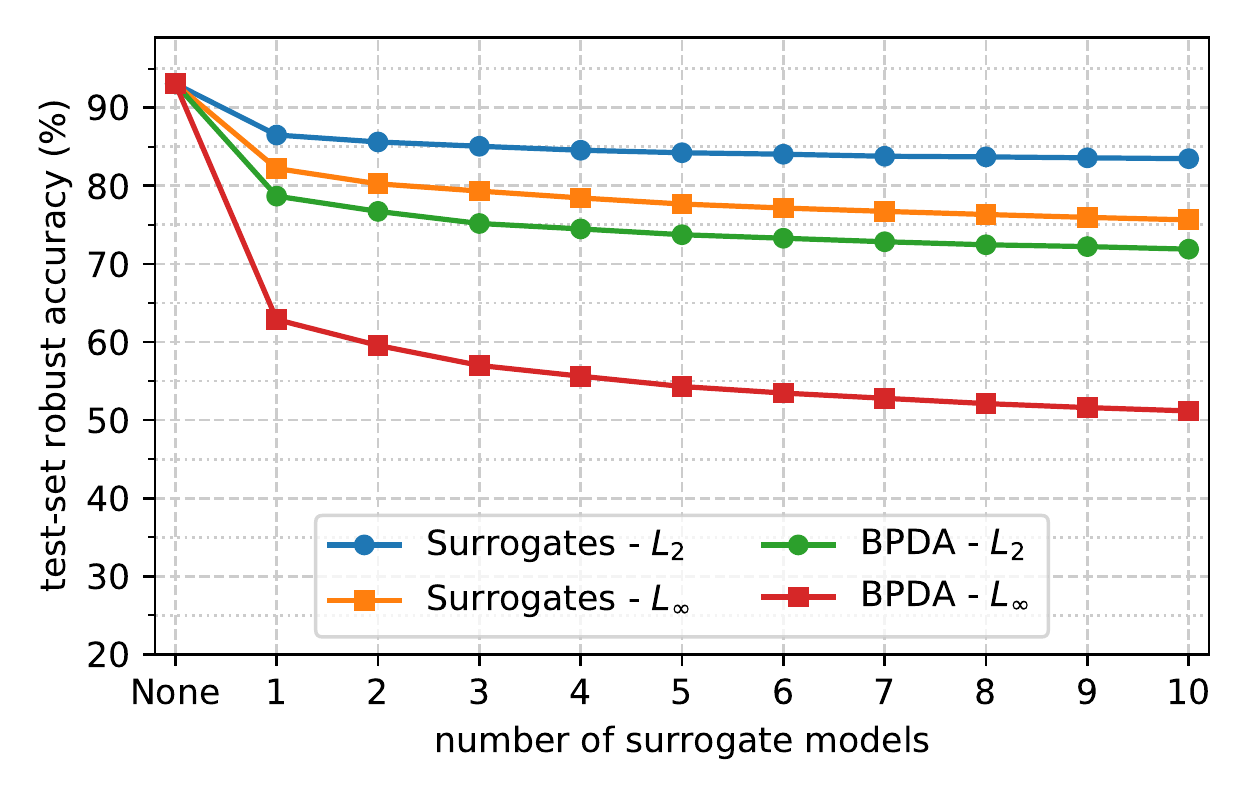}
\end{center}
\caption{Plots of CIFAR-10 test-set robust accuracy estimates (\bestknown) vs. number of attacker-trained surrogate models. We also plot the clean accuracy of $93.08\%$ for reference.\label{figs:acc_vs_models}}
\end{figure}
While we could not find a useful surrogate transformation by random guessing, it is still possible that transformations with different keys preserve largely the same input information. So, the second option of an attacker is to check whether decision boundaries of models defended with LINAC and different keys are in fact very similar, which would enable powerful transfer attacks from such \emph{surrogate models}. To this end, $10$ independent models defended with LINAC were trained from scratch, each using a different key chosen by the attacker. We used the most promising $10$ keys from the brute-force key attack for this purpose.
In Figure~\ref{figs:acc_vs_models} we report \bestknown robust accuracy  plotted against the number of surrogate models used in these joint attacks, and we aggregate results over all $10$ attacking models in the fourth column of Table~\ref{table:black-box}. However, this attack vector has limited success. Under transfer attacks with such surrogate models, the robust accuracy of our defended classifier appears to be high. While PGD and MT may fail due to vanishing or exploding gradients~\citep{Athalye2018obfuscated}, Square is a gradient-free attack, and does not suffer from such issues. Robust accuracy estimates according to Square are higher than $83\%$ against any individual surrogate model, irrespective of perturbation norm. A complete breakdown of results is given in Table~\ref{table:standard} of Appendix~\ref{sec:surrogates}.
Attacking with all $10$ surrogate models together, robust accuracy to Square is still higher that $75\%$, and the estimate is not improved by further aggregating over attacks. This evidence further support the hypothesis that decision boundaries of classifiers defended with LINAC depend on their respective keys, and may differ enough across keys to hinder transfer attacks with surrogates. Investing an order of magnitude more computation into such attacks leads to modest reductions in apparent robustness.

Lastly, an attacker may strive to employ BPDA, one of the most effective and general strategies against defences using nuisance transformations. BPDA attacks require: (1)~the ability to compute the exact forward transformation and (2)~finding a usefully differentiable approximation to the said transformation for use in the backwards pass of gradient-based attacks. In many cases this would be enough to allow the attacker to compute adversarial examples, perhaps at a somewhat higher computational cost~\citep{Athalye2018obfuscated, Tramer2020adaptive}.

Our LINAC defence presents further challenges by design. Exact forward computations (model inferences) require the \privatekey. An attacker cannot exactly compute the input transformation even for training set images, e.g. in order for some differentiable parametric approximation to be learned in a supervised fashion. Furthermore, surrogate models defended using LINAC and attacker chosen keys do not appear to be usefully differentiable, as suggested by results in  Table~\ref{table:black-box}. Nevertheless, an attacker could still hope that our defence ``filters out'' information in a largely key-agnostic manner, and that the choice of implicit network representation layer is not essential. Hence, they have the option of modifying LINAC to output activations of the last, rather than the middle layer of the implicit network. This amounts to reducing LINAC to an approximate reconstruction of the original signal. While such surrogate models with attacker chosen keys would still have to be trained for the purpose, they would be vulnerable to strong BPDA attacks, which may transfer well to our defended classifier. Apparent robustness estimates according to such transfer BPDA attacks are plotted in Figure~\ref{figs:acc_vs_models} as a function of the number of surrogate models used jointly in the attack. In the fifth column of Table~\ref{table:black-box} we provide aggregate apparent robust accuracies using $10$ such surrogates, showing that transfer BPDA attacks are more successful than previous attempts; any such reconstruction-based surrogate model can be used to reveal that the robust accuracy of our defended classifier cannot be higher than $65\%$, particularly with standard $L_{\infty}$ multi-targeted (MT) attacks (see Table~\ref{table:bpda} in Appendix~\ref{sec:surrogates} for a detailed breakdown of results). Interestingly, when $10$ surrogate models are used together, $L_{\infty}$ robust accuracy estimates drop to $51\%$. The reduction is less severe in standard $L_2$ attacks, where accuracy against all surrogates appears to be still over $71\%$. These results confirm that the BPDA strategy is a valuable tool for investigating the robustness of a wide range of defences, even when its assumptions are not fully met.

\begin{table}[t]
\centering%
\resizebox{\linewidth}{!}{%
\begin{tabular}{|L{.5cm}|L{1.75cm}|L{2.75cm}|L{1.5cm}|L{1.5cm}|}
\hline
\multicolumn{2}{|c|}{}  & {\bf All Source Models}    & \multicolumn{2}{c|}{\bf Adaptive Attacks}   \\
\hline
{\scriptsize{\bf Norm}} & {\bf Attack}   &   {Transfer}     & {BPDA}    &  {PBA}    \\
\hline\hline
\multirow{6}{*}{$L_\infty$}
        &	\autoattack	        &   41.18   &	59.40   &   68.34   \\
        &	\mt     	        &   47.91   &	55.37   &   46.75   \\
        &	\pgd    	        &   41.22   &	56.00   &   44.05   \\
        &	\sqr	            &   49.76   &	69.14   &   48.59   \\
        \cline{2-5}
        &   \bestknown          &   37.04   &   51.17   &   35.32   \\
\hline \hline
\multirow{6}{*}{$L_2$}      
        &	\autoattack	        &	71.32   &	74.59   &   73.10   \\
        &	\mt     	        &	73.83   &	74.98   &   67.85   \\
        &	\pgd    	        &	70.90   &	75.00   &   66.93   \\
        &	\sqr	            &	77.68   &	83.26   &   74.70   \\
        \cline{2-5}
        &   \bestknown          &   68.41   &   71.89   &   61.23   \\
        \hline
\end{tabular}%
}%
\caption{CIFAR-10 test set robust accuracy ($\%$) of a single LINAC defended classifier w.r.t. a suite of $L_\infty$ and $L_2$ attacks, valid under our threat model, using different strategies such as transfer and adaptive attacks. Our novel PBA adaptive attacks are overall more effective that both transfer and BPDA attack strategies.\label{table:adaptive}}%
\end{table}

\subsection{Transfer Attacks with Nominal and Adversarially Trained Source Models}
Since our defended classifier is not adversarially trained, one could assume that its decision boundaries may be similar to those of a nominal, undefended classifier. We show in the first column of Table~\ref{table:black-box} that transfer attacks with a nominally trained source model have limited success, especially considering that such undefended classifiers have below chance robust accuracies according to the very same evaluations. 

Another possibility is that that our defended model may be susceptible to the promising attack directions to which adversarially trained robust classifiers are vulnerable. We report in the second and third columns of Table~\ref{table:black-box} that this is indeed the case to some extent. Of all adversaries considered thus far, a robust model adversarially trained to tolerate  perturbations of up to size $\epsilon = 0.5$ in $L_2$ norm leads to the most effective transfer attacks. This holds to a lesser degree for an adversarially trained model with perturbations of size $\epsilon = 8/255$ in $L_\infty$ norm. Despite the success of evaluations using the former source model, no one attack method comes close to the effectiveness of the joint strategy, reported as \bestknown robust accuracy.

Furthermore, it is important to note that ensemble transfer attacks are much stronger than those computed with any given source model. Aggregated over four attack types and $23$ different source models, the robust accuracy of our LINAC defended classifier is revealed to be at most half of what initial results suggested  according to aggregate $L_\infty$ evaluations; this does not appear to be the case for $L_2$ attacks, however, which continue to be substantially hindered by LINAC. Robust accuracy could still be above $68\%$ according to the latter attack type, even in aggregate. In order to better characterise the implications of LINAC we make use of novel adaptive attacks in the following subsection.

\subsection{PBA Attacks Against LINAC}

Thus far we have shown that strong transfer attacks can be performed by using an ensemble of diverse source models to compute adversarial perturbations over many repeated trials. While ultimately more reliable, this is a cumbersome evaluation protocol, requiring two order of magnitude more computation than standard evaluations. 

In Section~\ref{subsec:PBA} we have introduced PBA, an attack strategy purposefully designed to be effective against input transformations (or network modules) which deny both inference and gradient computations, despite classifier parameters, training loss and dataset being available to the attacker. Following this novel strategy we successfully trained a parametric bypass approximation (PBA) of the LINAC transform and its associated \emph{defended classifier}. Intriguingly, the decision boundaries of the resulting \emph{bypass classifier} generalise very well. Accuracy on clean test data is $95.35\%$. Furthermore, the \emph{bypass classifier} can be readily shown to have $0\%$ robust accuracy using PGD attacks. This indicates that any apparent robustness in evaluations can be largely attributed to the LINAC transform successfully hindering attacks, since the decision boundaries of our \emph{defended classifier} are susceptible to adversarial perturbations, and hence cannot be considered to add any inherent robustness by themselves.

\begin{figure*}[h]
\centering
\begin{tabular}{cc}
\tabcaption & \includegraphics[align=c,width=0.93\textwidth]{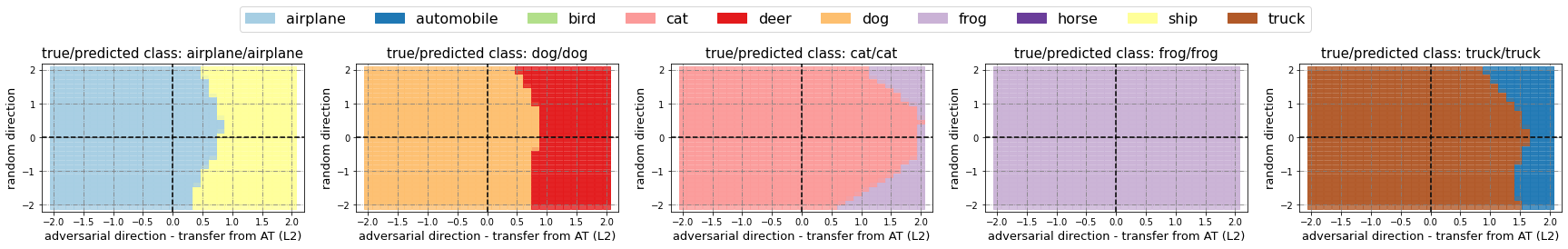} \\
\tabcaption & \includegraphics[align=c,width=0.93\textwidth]{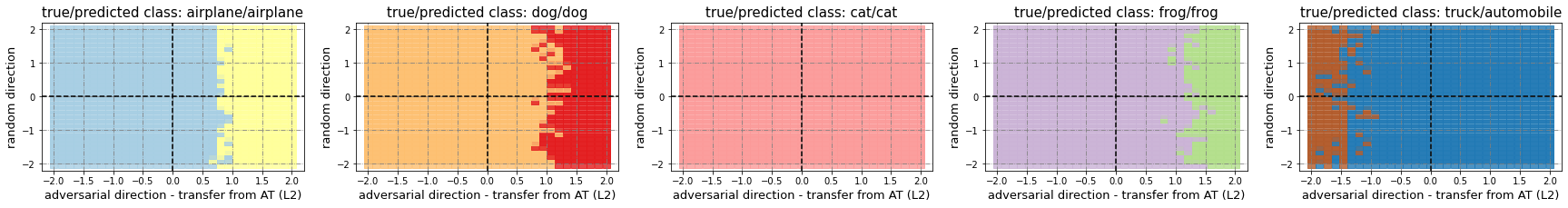}  \\ 
\tabcaption & \includegraphics[align=c,width=0.93\textwidth]{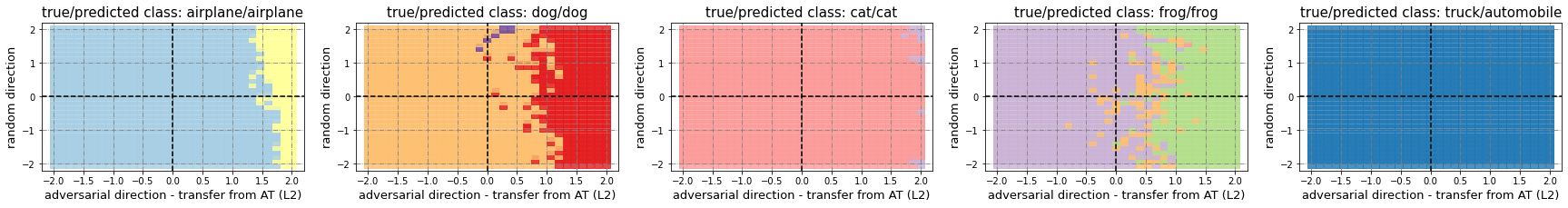}  \\ 
\tabcaption & \includegraphics[align=c,width=0.93\textwidth]{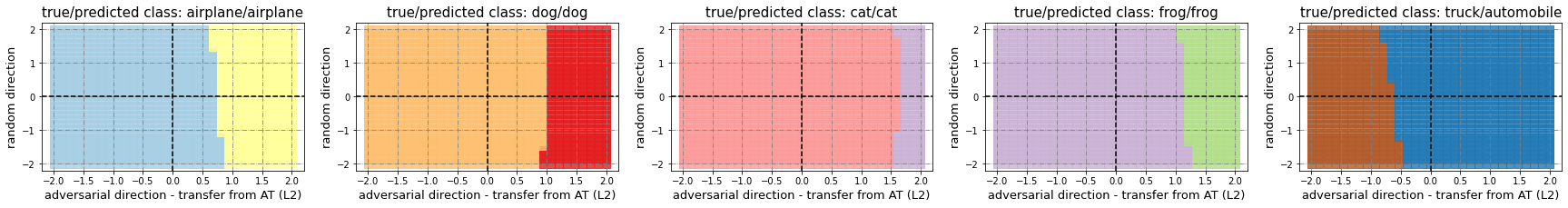}  \\ 
\tabcaption & \includegraphics[align=c,width=0.93\textwidth]{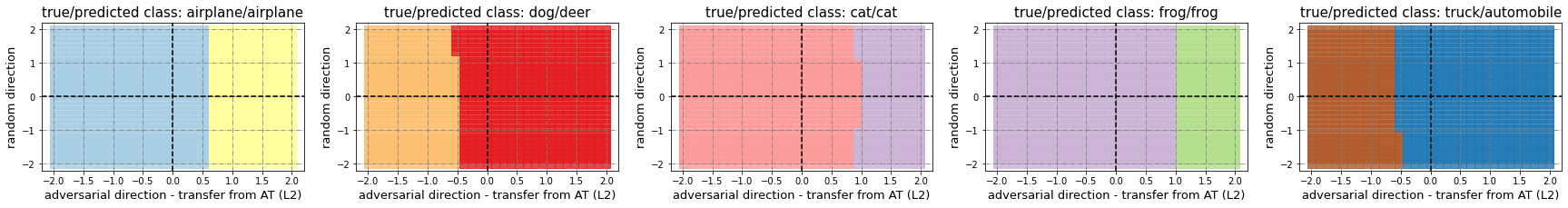}  \\ 
\end{tabular}
\caption{Decision boundaries of five different classifiers (rows) around the same five randomly chosen test examples (columns), plotted along their respective adversarial directions according to the AT ($L_2$) model (horizontal), and the same random direction (vertical): (A) An undefended, nominally trained CIFAR-10 classifier; (B) LINAC defended classifiers using a random key; (C) LINAC defended classifiers using the \privatekey; this is the model we evaluated throughout the submission; (D) The by-pass classifier resulting from our novel PBA attack on model (C); (E) An adversarially trained classifier, AT ($L_2$) in the main text, used to generate transfer attacks. We observe that boundaries of nominal model (A) are different from those of LINAC (B) \& (C); LINAC decision boundaries seem less smooth compared to other models; as suspected boundaries appear different across keys (B) vs. (C), which corroborates our observations of robustness to transfer attacks with surrogates. The adversarially trained model (E) is robust to the vertical dimension (random noise); LINAC models (B) \& (C) also appear less sensitive to random noise compared the nominal model (A). PBA by-pass classifier (D) boundaries are much smoother and different from the true boundaries of the attacked model (C), which may explain why LINAC withstands the novel attack in many cases. Notice that PBA approximated boundaries (D) can be both closer and farther away from test examples compared to the true model's (C), which makes it less clear how useful such approximations are for future attacks on LINAC.\label{figs:variants}}
\end{figure*}

In Table~\ref{table:adaptive} we show that standard attacks using the trained PBA mapping against our LINAC defended classifier are even more effective than BPDA attacks using $10$ source models. Interestingly, PBA almost uniformly leads to more effective attacks, regardless of strategy. PGD attacks using PBA give the most accurate picture of robustness of all strategies, suggesting that the matter of \emph{obfuscated gradients} is largely mitigated by our novel strategy. Aggregated over different attack types, PBA is the most effective and efficient evaluation strategy which does not make use of the \privatekey, and hence is valid under the adopted threat model. Based on these evaluations alone, one may conclude that robust accuracy was over $35\%$ under attacks of size at most $\epsilon = 8/255$ in $L_\infty$ norm, and over $61\%$ for attacks of size $\epsilon = 0.5$ in $L_2$ norm. The apparent robustness difference between $L_\infty$ and $L_2$ attacks persists, suggesting that LINAC primarily hinders the latter type of attacks.

\subsection{Towards Explaining the Apparent Robustness}
{\bf Decision boundary inspection.} We plotted decision boundaries of several classifiers around five randomly chosen test examples in Figure~\ref{figs:variants}. All boundary plots are centred on test examples (columns), use appropriate adversarial directions as the horizontal dimension, and a random direction as the vertical. As expected, we observe differences between LINAC defended classifiers which use different keys. Furthermore, we found that LINAC boundaries can be more ``complicated'' relative to those of other models, which may explain why PBA attacks are not completely effective.

{\bf RGB Reconstruction vs.~Lossy Encodings.} Setting the representation layer index $K = L$ renders our LINAC transform into an approximate RGB input reconstruction, since $L$ is the index of the implicit network output layer. We confirmed that setting $K = L = 5$ and $N=100$ epochs offers no robustness, since the resulting reconstructions are precise and BPDA attacks are successful. Clean accuracy was $96.91\%$, virtually matching that of a nominally trained classifier. Hence, any apparent robustness must be due to the number of INR fitting epochs $N$, and/or the choice of representation layer index $K$. Intuitively, both hyper-parameters control how ``lossy'' our transformation is.

Naturally, we were interested in reducing the computational overhead of LINAC. Aiming to match the clean accuracy of state-of-the-art adversarially trained robust classifiers, specifically $93\%$ \citep{Rebuffi2021fixing}, we empirically chose $N=10$ epochs as a trade-off between speed and clean accuracy. The activation coding layer index $K=2$ out of $L=5$ hidden layers was chosen according to the same principle, as the lowest level representation which did not reduce clean accuracy below the target threshold. We further characterise and illustrate our LINAC transform in Appendix~\ref{sec:lossy_encoding}.

{\bf Performance Considerations.} LINAC is as expensive as inference with a WideResNet-70-16 \citep{ZagoruykoKomodakis2016wrn} on CIFAR-$10$ images. This cost is dominated by the fitting of INRs.
 It could be reduced with an adaptive form of ``early stopping'' based on loss values, or by leveraging advances in INR research (e.g.~\citet{Sitzmann2020implicit}). We leave these investigations, and scaling LINAC to larger images, for future work.

{\bf Sensitivity Analyses.} The apparent robustness of LINAC defended classifiers is largely insensitive to the number of hidden layers $L \ge 3$ of the implicit MLP, as well as the number of features $F \ge 3$ in its positional input encoding, hence we relegated the sensitivity analyses to Appendix~\ref{sec:sensitivity_analysis}.

\begin{table}[t]
\centering

\resizebox{\linewidth}{!}{%
\begin{tabular}{|L{0.6cm}|L{1cm}|L{1cm}L{1cm}L{1cm}L{1cm}L{1.2cm}|L{1cm}|}
\hline
\multicolumn{2}{|L{1.6cm}|}{}  & {\bf Full PCA}  & {\bf Block PCA} & {\bf JPEG (23)} & {\bf JPEG (10)} & {\bf Block Pixel Shuffle}    &  {\bf LINAC (Ours)}\\
\hline
\multicolumn{2}{|L{1.6cm}|}{\it Clean Accuracy:}   &   96.10   &   96.39   &   88.15   &   81.17   &   96.98   &   93.08   \\
\hline\hline
{\scriptsize{\bf Norm}} & {\bf Attack} & {Standard} & {Standard} & {BPDA} & {BPDA} & {PBA}  & {PBA} \\
\hline\hline
\multirow{6}{*}{$L_\infty$}
        &	\autoattack	        &   0.00    &   0.00    &   11.90   &   32.58   &   0.18    &   68.34   \\
        &	\mt     	        &   0.00    &   0.00    &   --      &   --      &   0.00    &   46.75   \\
        &	\pgd    	        &   0.00    &   0.00    &   17.49   &   27.48   &   0.00    &   44.05   \\
        &	\sqr	            &   0.00    &   0.00    &   5.36    &   6.34    &   0.00    &   48.59   \\
        \cline{2-8}
        &   \bestknown          &   0.00    &   0.00    &   0.61    &   2.26    &   0.00    &   35.32   \\
\hline\hline
\multirow{6}{*}{$L_2$}      
        &	\autoattack	        &    0.00   &    0.06   &   62.95   &   62.98   &    0.02   &   73.10   \\
        &	\mt     	        &    0.03   &    0.00   &    --     &   --      &    0.00   &   67.85   \\
        &	\pgd    	        &    0.41   &    0.17   &   60.37   &   60.38   &    0.02   &   66.93   \\
        &	\sqr	            &   12.85   &   11.66   &   28.33   &   21.92   &    6.13   &   74.70   \\
        \cline{2-8}
        &   \bestknown          &   0.02    &   0.00    &   14.94   &   14.56   &    0.00   &   61.23   \\
\hline
\end{tabular}%
}
\caption{CIFAR-10 test set robust accuracy ($\%$) of several classifiers defended using related input transformations according to evaluations using adversarial perturbations bounded in $L_\infty$ and $L_2$ norms. Reporting results of strongest known attack strategy for each method, valid according to its own threat model.\label{table:baselines}}
\end{table}

\subsection{PBA Beyond LINAC and Methodology Validation}

We show in the one-but-last column of  Table~\ref{table:baselines} that PBA successfully and completely invalidates the Block Pixel Shuffle approach of \citet{Aprilpyone2021block}, despite its good reported robustness against all attacks. We further investigate using adversarially trained source models, see full results in Table~\ref{table:blockshuffle} of Appendix~\ref{sec:surrogates}. In summary, our analysis confirms that the apparent robust accuracy of Block Pixel Shuffle according to valid attacks bounded in $L_2$ norm remains high at $69\%$.  Hence, PBA is indeed the only known valid attack on this defence which is completely successful.

Finally, we validate our evaluation methodology by testing its effectiveness against similar defences. We perform the same evaluations on the Principal Component Analysis (PCA) based defence of \citet{Shaham2018defending}, and the JPEG-based defences of \citet{Das2017keeping, Das2018shield, Guo2018countering}. In Table~\ref{table:baselines} we report the \bestknown robust accuracies of these defences according to our evaluation methodology, which are directly comparable with our reported LINAC results. We observe that LINAC successfully hinders much stronger attacks than these alternative strategies.

\section{Conclusions}
In this work we introduce LINAC, a novel key-based defence using implicit neural representations, and demonstrate its effectiveness for hindering standard adversarial attacks on CIFAR-$10$ classifiers.
We systematically attempt to circumvent our defence by adapting a host of widely used attacks from the literature, including transfer and adaptive attacks, but LINAC maintains strong apparent robustness.
%
Consequently, we challenge LINAC by introducing a novel adaptive attack strategy (PBA) which is indeed more successful at discovering adversarial examples. We also show that PBA can be used to completely invalidate an existing key-based defence. 
%
These are some of the latest attempts to leverage computational hardness for adversarial robustness, and successful PBA attacks on existing methods  enable further progress.

\bibliography{main}
\bibliographystyle{icml2022}

\newpage
\appendix
\onecolumn

\addtolength{\textfloatsep}{0.25in}  

\section{LINAC Implementation Details}
\label{sec:impl_details}
\subsection{Implicit Neural Representations}

\subsubsection{Random number generation for INRs}

Our LINAC defence is fully deterministic by design. We used a random $64$-bit signed integer as the \privatekey, which seeded the state of the pseudo-random number generator in JAX \citep{Frostig2018compiling, jax2018github}. The precise value of the \privatekey used to train the defended model evaluated throughout this work was: $-2314326399425823309$. It was itself selected randomly, by initialising the random number generator of the NumPy library \citep{Harris2020array} with seed $42$ and using the first $\text{int}64$ integer.


\subsubsection{Input and Output Encodings}
Following \citet{Mildenhall2020nerf} we use a positional encoding of pixel coordinates to a higher dimensional space to better capture higher-frequency information. Each pixel coordinate $d$ is normalised to $[-1, 1]$ and transformed as follows:
\begin{equation}
\begin{multlined}
\gamma(d) = [\mathrm{sin}(2^0 \pi d), \mathrm{cos}(2^0 \pi d), \mathrm{sin}(2^1 \pi d), \mathrm{cos}(2^1 \pi d), \dots,  \mathrm{sin}(2^{F-1} \pi d), \mathrm{cos}(2^{F-1} \pi d)]
\end{multlined}
\end{equation}
We used $F=5$ frequencies in all our experiments and a $L=5$ hidden layer MLP with $H=256$ units per layer and ReLU non-linearities. Activations in the middle hidden layer were used for computing the LINAC transform, hence $K=2$.

As per standard practice for CIFAR-$10$ classification, pixel colour intensities were scaled to have $0$ mean across the training dataset and each colour channel separately. Intensities were then standardised to $1$ standard deviation across the training dataset, independently across channels.

\subsubsection{Fitting}\label{app:fitting}
Fitting the parameters of the implicit neural network was done using Adam~\cite{Kingma2015adam}, with default parameters and a learning rate $\mu = 0.001$. We used mini-batches with $M=32$ random pixels and trained for $N=10$ epochs. An epoch constitutes a pass through the entire set of pixels in the input image with dimensions $I \times J \times C = 32 \times 32 \times 3$ in random order. The total number of optimisation steps performed was $320$. A cosine learning rate decay schedule was used for better convergence, with the minimum value of the multiplier $\alpha = 0.0001$ \citep{Loshchilov2016SGRD}.

\subsubsection{Computational and Memory Requirements}\label{app:compreq}
The LINAC transform's computational complexity scales with the number of pixels ($I \cdot J$) of the input image and the number of epochs through the pixels ($N$). It takes $I\cdot J \cdot N$ backward passes through the implicit network $\mlp$ to fit its parameters $\mlpparam$. LINAC's memory complexity is dominated by the number of parameters of the INR ($|\mlpparam|$). Empirically, the LINAC transform is itself as expensive as inference with a WideResNet-70-16 model \citep{ZagoruykoKomodakis2016wrn} on CIFAR-$10$ images.


\subsection{Defended Classifiers}
Since the proposed input transformation preserves spatial structure, we perform image classification using transformed inputs in an identical manner as with RGB colour images, except for the higher number of channels of transformed inputs. Hence, we employ a standard classification pipeline following \citep{ZagoruykoKomodakis2016wrn}, using a WideResNet-70-16 classifier. We reiterate that our proposed transformation changes the number of input channels, but not the spatial dimensions. Hence, small differences between our models and other WideResNet-70-16 results reported in the literature could conceivably appear only due to different numbers of input channels. However, practically this leads to less than a $0.2\%$ increase in the total number of model parameters, limited to the first convolutional layer, which uses filters with $256$ channels instead of $3$.

We used the Swish activation function proposed by \citet{Ramachandran2017searching} for all the classifiers. Training was performed with Nesterov Momentum SGD \citep{Tieleman2012lecture} $m=0.9$, using mini-batches of size $1024$, for a total of $1000$ epochs, or $48880$ parameter updates. The initial learning rate was $\mu=0.4$, reduced by a factor of 10 four times, at epochs: $650$, $800$, $900$ and $950$. We performed a hyper-parameter sweep over the weight-decay scale with the following grid: $\{0., 0.0001, 0.0005, 0.0010\}$. 
We maintain an exponential moving average of classifier parameters (with a decay rate of $r=0.995$); we report accuracies using the final average of classifier parameters.

\subsubsection{Performance Considerations}
\label{sec:performance_considerations}

We use the CutMix data augmentation strategy of \citet{Yun2019cutmix} directly on RGB images from the training set of CIFAR-10, prior to transforming them with LINAC. This has an impact on computational considerations, since pre-computing the transformed dataset offline in order to save training time becomes more challenging. For ease of prototyping we chose to implement LINAC as a preprocessing layer, which could have an impact on training time if used naively, but not if the transformation is applied asynchronously on the buffer of data feeding the device used for model training. We also found empirically that the proposed transformation renders itself to very effective parallelisation using modern SIMD devices, despite the fact that there is no parameter sharing between implicit models of different inputs; this is likely due to the ability of modern libraries such as JAX \citep{Frostig2018compiling} to vectorise operations across tensors holding parameters for many distinct neural networks.

It is important to note that inference and training costs of defended classifiers are roughly double those of the nominal classifier. Hence, the LINAC transform has comparable cost to inference with a WideResNet-70-16 model.

\section{Evaluation Details}
\subsection{Attacks with Surrogate Models}
\label{sec:surrogates}

We provide a breakdown of evaluations using surrogate models initially reported in Section~\ref{sec:results}. We report the best $10$ keys from the brute-force attack on the \privatekey in Table~\ref{table:best_keys}. These keys were also used to train surrogate models defended with LINAC for use in transfer attacks, see Table~\ref{table:standard} for complete results. Reconstruction-based surrogate models defended with modified LINAC, and using the same $10$ best-guess attacker keys, were used to perform BPDA transfer attacks, reported in Table~\ref{table:bpda}.

\begin{table}[h]
\centering
\begin{tabular}{|c|c|c|}
\hline
Position& Clean Test Accuracy $(\%)$   &   Attacker Key \\
\hline\hline
1       &   57.00    &   1383227977468296715   \\
\hline
2       &   55.00    &   -3328443931658504707  \\
\hline
3       &   55.00    &   -127094507362684985   \\
\hline
4       &   55.00    &   -7808219206569127925  \\
\hline
5       &   55.00    &   -8772667224621836765  \\
\hline
6       &   55.00    &   -70640792831170485    \\
\hline
7       &   54.00    &   8263151932495004089   \\
\hline
8       &   54.00    &   -4594861196100637268  \\
\hline
9       &   54.00    &   -6520968232434877967  \\
\hline
10      &   54.00    &   -8722766234183220599  \\
\hline
\end{tabular}
\caption{Top $10$ keys in brute-force key attack, also used to train surrogate models.\label{table:best_keys}}
\end{table}

\begin{table}[h]
\centering
\resizebox{.95\textwidth}{!}{%
\begin{tabular}{|c|c|cccccccccc|c|}
\hline
\multicolumn{2}{|c|}{\bf LINAC Defence}   & \multicolumn{10}{c|}{\bf Defended Surrogate Source Models (Attacker Keys)}  & \multicolumn{1}{c|}{\bestadversary}
\\ \hline
\multicolumn{1}{|c|}{\bf Norm} & {\bf Attack Name}    & {\bf $key_1$}  & {\bf $key_2$} & {\bf $key_3$} & {\bf $key_4$} & {\bf $key_5$} & {\bf $key_6$} & {\bf $key_7$} & {\bf $key_8$} & {\bf $key_9$} & {\bf $key_{10}$}  & (against all models)    \\
\hline\hline
\multirow{5}{*}{$L_\infty$}	
	&	\autoattack	        &	93.05	&	92.99	&	93.06	&	93.00	&	93.01	&	93.05	&	93.00	&	93.00	&	93.15	&	93.21	&	84.00	\\
	&	\mt     	        &	89.70	&	89.48	&	89.63	&	89.54	&	89.59	&	89.55	&	89.36	&	89.58	&	89.69	&	89.47	&	85.70	\\
	&	\pgd    	        &	88.05	&	88.19	&	88.20	&	88.23	&	88.24	&	88.17	&	88.06	&	88.20	&	88.29	&	88.10	&	87.32	\\
	&	\sqr	            &	83.37	&	83.34	&	83.60	&	83.43	&	83.06	&	83.38	&	83.41	&	83.30	&	83.36	&	83.21	&	75.91	\\
	\cline{2-13}
	&   \bestknown          &   82.22   &   82.10   &   82.43   &   82.26   &   81.93   &   82.33   &   82.24   &   82.10   &   82.32   &   82.17   &   75.64   \\
\hline\hline
\multirow{5}{*}{$L_2$}
	&	\autoattack	        &	91.20	&	91.15	&	91.24	&	91.17	&	91.18	&	91.18	&	91.20	&	91.20	&	91.21	&	91.16	&	88.27	\\
	&	\mt     	        &	90.29	&	90.47	&	90.30	&	90.50	&	90.54	&	90.37	&	90.20	&	90.32	&	90.56	&	90.25	&	87.31	\\
	&	\pgd    	        &	88.99	&	89.00	&	88.92	&	88.93	&	88.93	&	89.08	&	88.97	&	88.98	&	88.95	&	89.03	&	88.36	\\
	&	\sqr	            &	87.84	&	87.83	&	87.77	&	88.15	&	87.88	&	88.11	&	88.19	&	87.94	&	87.85	&	88.12	&	84.08	\\
	\cline{2-13}
	&   \bestknown          &   86.51   &   86.41   &   86.59   &   86.42   &   86.33   &   86.62   &   86.50   &   86.58   &   86.58   &   86.53   &   83.48   \\
	 \hline
\end{tabular}%
}
\caption{CIFAR-$10$ test set robust accuracy ($\%$) of a single LINAC defended classifier w.r.t. a suite of $L_\infty$ and $L_2$ transfer attacks, valid under our threat model, using surrogate classifiers defended with LINAC, but trained with attacker-chosen keys. The clean accuracy of our defended classifier is $93.08\%$.\label{table:standard}}
\end{table}

\begin{table}[h]
\centering
\resizebox{.95\textwidth}{!}{%
\begin{tabular}{|c|c|cccccccccc|c|}
\hline
\multicolumn{2}{|c|}{\bf LINAC Defence}   & \multicolumn{10}{c|}{\bf Reconstruction-Based Surrogate Source Models Using Attacker Keys}  & {\bestadversary}
\\ \hline
\multicolumn{1}{|c|}{\bf Norm} & \multicolumn{1}{c|}{\bf Attack Name}    & {\bf $key_1$}  & {\bf $key_2$} & {\bf $key_3$} & {\bf $key_4$} & {\bf $key_5$} & {\bf $key_6$} & {\bf $key_7$} & {\bf $key_8$} & {\bf $key_9$} & {\bf $key_{10}$}  & (against all models)
\\ \hline
\hline
\multirow{5}{*}{$L_\infty$}	
	&	\autoattack	        &	91.47	&	91.24	&	91.21	&	91.31	&	91.32	&	91.35	&	91.31	&	91.63	&	91.33	&	91.46	&	59.40   \\
	&	\mt     	        &	68.35	&	69.13	&	67.71	&	68.63	&	67.53	&	69.07	&	68.81	&	67.98	&	68.40	&	69.24	&	55.37   \\
	&	\pgd    	        &	69.35	&	71.57	&	69.15	&	70.55	&	69.30	&	70.46	&	69.67	&	69.48	&	69.97	&	70.85	&	56.00   \\
	&	\sqr	            &	82.91	&	82.98	&	82.47	&	82.92	&	82.66	&	82.80	&	82.84	&	82.91	&	82.91	&	82.93	&	69.14   \\
	\cline{2-13}
	&   \bestknown          &   62.87   &   64.59   &   62.63   &   63.65   &   62.40   &   64.02   &   63.54   &   62.93   &   63.40   &   64.15   &   51.17   \\
	\hline
\hline
\multirow{5}{*}{$L_2$}
	&	\autoattack	        &	85.94	&	86.29	&	85.95	&	86.51	&	85.52	&	86.16	&	86.28	&	86.34	&	86.46	&	86.27	&	74.59   \\
	&	\mt     	        &	82.12	&	82.27	&	81.91	&	82.13	&	81.78	&	82.42	&	82.24	&	82.07	&	82.21	&	81.96	&	74.98   \\
	&	\pgd    	        &	82.23	&	82.77	&	82.20	&	82.60	&	81.89	&	82.33	&	82.53	&	82.38	&	82.69	&	82.68	&	75.00   \\
	&	\sqr	            &	87.30	&	87.34	&	87.22	&	87.40	&	87.35	&	87.44	&	87.52	&	87.21	&	87.26	&	87.46	&	83.26   \\
	\cline{2-13}
	&   \bestknown          &   78.68   &   79.15   &   78.36   &   78.93   &   78.46   &   79.08   &   78.92   &   78.83   &   78.80   &   78.46   &   71.89   \\
	\hline
\end{tabular}%
}
\caption{CIFAR-$10$ test set robust accuracy ($\%$) of a single LINAC defended classifier according to a suite of $L_\infty$ and $L_2$ BPDA attacks, valid under our threat model, using reconstruction-based surrogate classifiers.\label{table:bpda}}
\end{table}

\subsection{Transfer Attacks with Adversarially Trained Models}
For mounting transfer attacks we have taken adversarially trained models from previous work~\cite{Rebuffi2021fixing}, with checkpoints available online\footnote{\url{https://github.com/deepmind/deepmind-research/tree/master/adversarial_robustness}}. These models have been adversarially trained on CIFAR-10 using additional synthetic generated data and CutMix data augmentation. To mount transfer attacks we use the WideResNet-106-16 model (trained to defend against $L_{\infty}$ norm-bounded perturbations of size $\epsilon=8/255$) and the WideResNet-70-16 model (trained to defend against $L_2$ norm-bounded perturbations of size $\epsilon=0.5$).

\subsection{PBA Implementation Details}

\subsubsection{PBA for LINAC}
We used a single convolutional layer ($k = 3 \times 3$)  with biases to implement  $h_{\psi}(x)$, the PBA of the nuisance transformation, mapping from the $3$ RGB channels of input images to the $H=256$ channels output by LINAC. 

The parameters $\psi$ of the bypass approximation were trained by minimising the cross-entropy loss on the CIFAR-$10$ training set using Momentum SGD with a learning rate $\mu = 0.1$. $100$ epochs sufficed to optimise PBA parameters, with four learning rate reductions by a factor of $0.1$ at epochs: $65$, $80$, $90$, $95$.

\subsubsection{PBA for Block Pixel Shuffle}

\begin{figure}[h]
\centering
\includegraphics[width=0.6\linewidth]{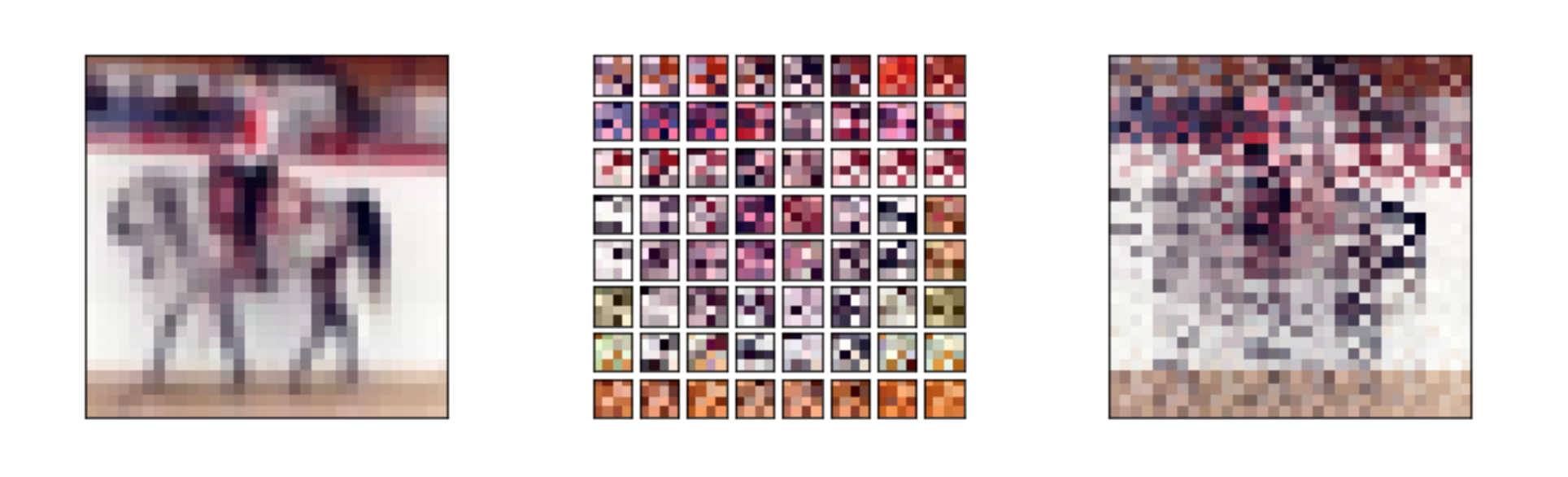}

\caption{Example of Pixel Block Shuffle transformation \citep{Aprilpyone2021block}. An original CIFAR-$10$ image (left) is split into a grid of $4 \times 4$ blocks of adjacent pixels, and the same random permutation is used to shuffle pixel positions within every block (middle). The transformed image is constructed by spatially concatenating the blocks according to their original positions in the grid (right). \label{figs:blockshuffle}}
\end{figure}

We implemented the Block Pixel Shuffle defence of \citet{Aprilpyone2021block} using blocks of size $4 \times 4$, as recommended in the original work. We used the same \privatekey value as that of our defended LINAC classifier. The \privatekey serves as the seed of a pseudo-random number generator, which is used to sample a permutation of all pixel positions in a block. The same permutation is applied to all blocks. We illustrate the transform in Figure~\ref{figs:blockshuffle}.

A classifier defended with Block Pixel Shuffle was trained with the same procedure as our defended LINAC classifier. We can report a clean CIFAR-$10$ test set accuracy of $97.03\%$, which is higher to that reported by \citet{Aprilpyone2021block}, but consistent with the superior CutMix \citep{Yun2019cutmix} data augmentation procedure we used for all defended classifiers.

According its own ``white-box'' threat model \citep{Aprilpyone2021block}, all the implementation details of the defence are known to an attacker except the \privatekey. We exploit the block structure and use a single linear layer without biases, and initialised with the identity mapping, to compute a parametric bypass approximation (PBA) for the this defence. We found that using a smaller initial learning rate $\mu = 0.001$ results in stable convergence. We used $300$ epochs to optimise PBA parameters, with four learning rate reductions by a factor of $0.1$ at epochs: $275$, $285$, $290$, $295$.

An extensive evaluation of the resulting defended classifier is given in Table~\ref{table:blockshuffle}. We find that transfer attacks which are agnostic to the defence can be more successful when adversarial examples are computed using robust source models, but one may infer some level of robustness. Using PBA attacks valid under the threat model (``white-box'') we successfully circumvent the defence, with a \bestknown CIFAR-$10$ robust test-set accuracy of $0\%$ under adversarial perturbations of size up to $\epsilon = 8 / 255$ in $L_\infty$ norm, and up to $\epsilon = 0.5$ in $L_2$ norm.

\begin{table}[h]
\centering
\begin{tabular}{|l|l|L{1.5cm}L{1.7cm}L{1.7cm}|L{1.5cm}|L{1.7cm}|}
\hline
\multicolumn{2}{|l|}{\bf Block Pixel Shuffle Defence}  & \multicolumn{3}{c|}{\bf Transfer Attack Source Models}  & {\bf Adaptive Attacks}  & {\bestadversary} \\
\hline
{\bf Norm} & {\bf Attack Name} &   {\bf Nominal Source} & {\bf Adversarial Training $(\mathbf{L_\infty})$}  & {\bf Adversarial Training $(\mathbf{L_2})$} & {\bf PBA}  & {\bf All Source Models}\\
\hline\hline
\multirow{5}{*}{$L_\infty$}
        &	\autoattack	        &	85.78	&	69.09	&	73.86	&	0.18	&   0.00    \\
        &	\mt     	        &	78.87	&	56.49	&	27.72	&	0.00	&   0.00    \\
        &	\pgd    	        &	69.17	&	39.05	&	31.19	&	0.00	&   0.00    \\
        &	\sqr	            &	69.16	&	46.25	&	42.63	&	0.00	&   0.00    \\
        \cline{2-7}
        &   \bestknown          &   60.65   &   30.61   &   21.17   &    0.00   &   0.00    \\
\hline\hline
\multirow{5}{*}{$L_2$}      
        &	\autoattack	        &	94.14	&	90.02	&	83.35	&	0.02	&   0.00    \\
        &	\mt     	        &	93.93	&	92.25	&	77.02	&	0.00	&   0.00    \\
        &	\pgd    	        &	92.92	&	87.54	&	77.80	&	0.02	&   0.02    \\
        &	\sqr	            &	91.69	&	88.80	&	84.88	&	6.13	&   6.13    \\
        \cline{2-7}
        &   \bestknown          &   90.41   &   85.29   &   69.00   &   0.00    &   0.00    \\
\hline
\end{tabular}

\caption{CIFAR-10 test set robust accuracy ($\%$) of Block Pixel Shuffle approach \citep{Aprilpyone2021block} against standard $L_\infty$ and $L_2$ bounded attacks using both transfer and our novel PBA strategy.\label{table:blockshuffle}}
\end{table}


\section{Sensitivity of LINAC to Hyper-Parameters}
\label{sec:sensitivity_analysis}

We performed sensitivity analyses of LINAC to its hyper-parameters. For efficiency reasons we report robust accuracies according to untargeted PGD attacks with $100$ steps and $10$ restarts, using an adversarially trained robust model ($L_2$) \citep{Rebuffi2021fixing} to generate adversarial perturbations.

\begin{figure}[h]
\centering
\begin{minipage}[h]{0.45\linewidth}
\centering
\includegraphics[width=0.95\linewidth]{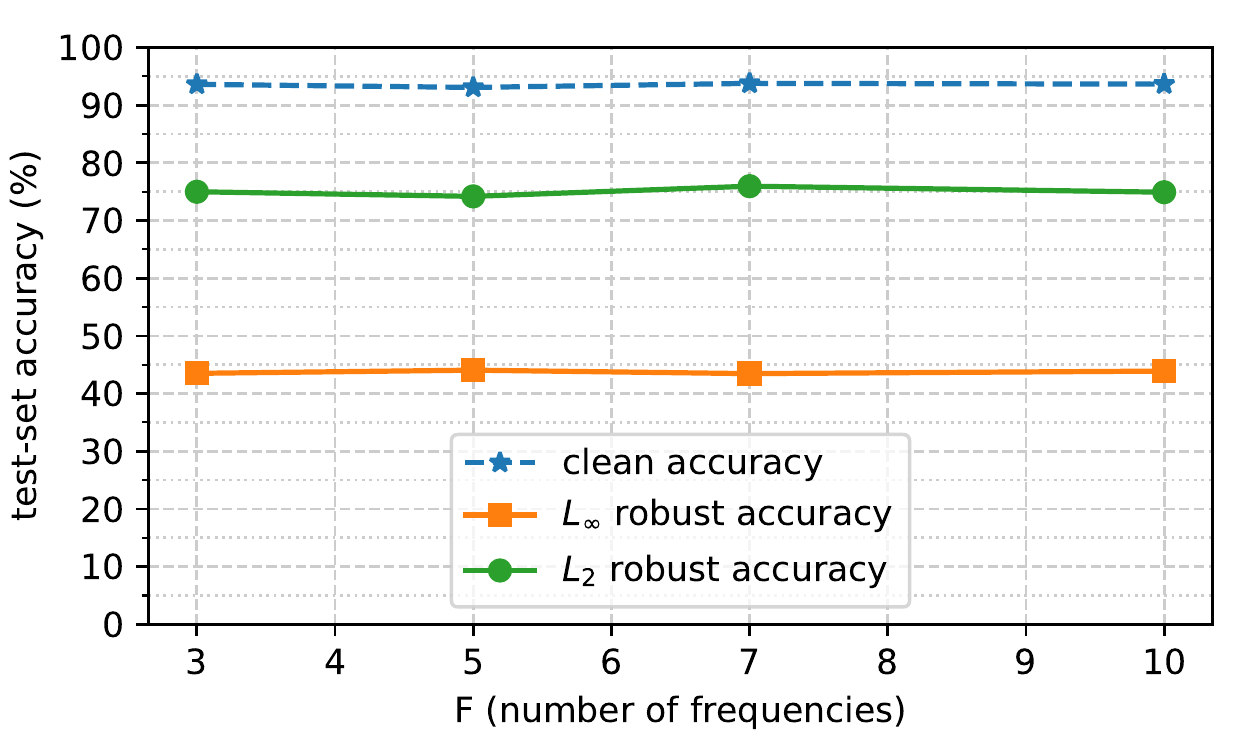}
\end{minipage}\qquad\begin{minipage}[b]{0.45\linewidth}
\centering
\resizebox{1\linewidth}{!}{%
\begin{tabular}{|l|l|l|l|l|l|l|l|}
\hline
\multicolumn{2}{|L{1.6cm}|}{}  &	$F=3$	&	$F=5$	&	$F=7$	&	$F=10$	\\
\hline
\multicolumn{2}{|L{1.6cm}|}{\it Clean Accuracy:}    &	93.61	&	93.08	&	93.78	&	93.65	\\
\hline\hline
$L_\infty$  &   \pgd                                &	43.51	&	44.06	&	43.46	&	43.90	\\
\hline\hline
        $L_2$       &   \pgd                        &	74.99	&	74.19	&	75.96	&	74.91	\\
\hline
\end{tabular}
}
\end{minipage}
\caption{CIFAR-$10$ test-set clean and robust accuracies under transfer attacks of LINAC defended classifiers with different numbers of positional encoding frequencies $F$, keeping all other hyper-parameters constant.\label{figs:linac_sensitivity_F}}
\end{figure}

In Figure~\ref{figs:linac_sensitivity_F}  we provide a sensitivity analysis across the number of frequencies $F$ used for positional encoding \citep{Mildenhall2020nerf}, keeping all other hyper-parameters the same. Note that we used $F=5$ for our defended classifier evaluated in the main paper.

\begin{figure}[h]
\centering
\begin{minipage}[h]{0.45\linewidth}
\centering
\includegraphics[width=0.95\linewidth]{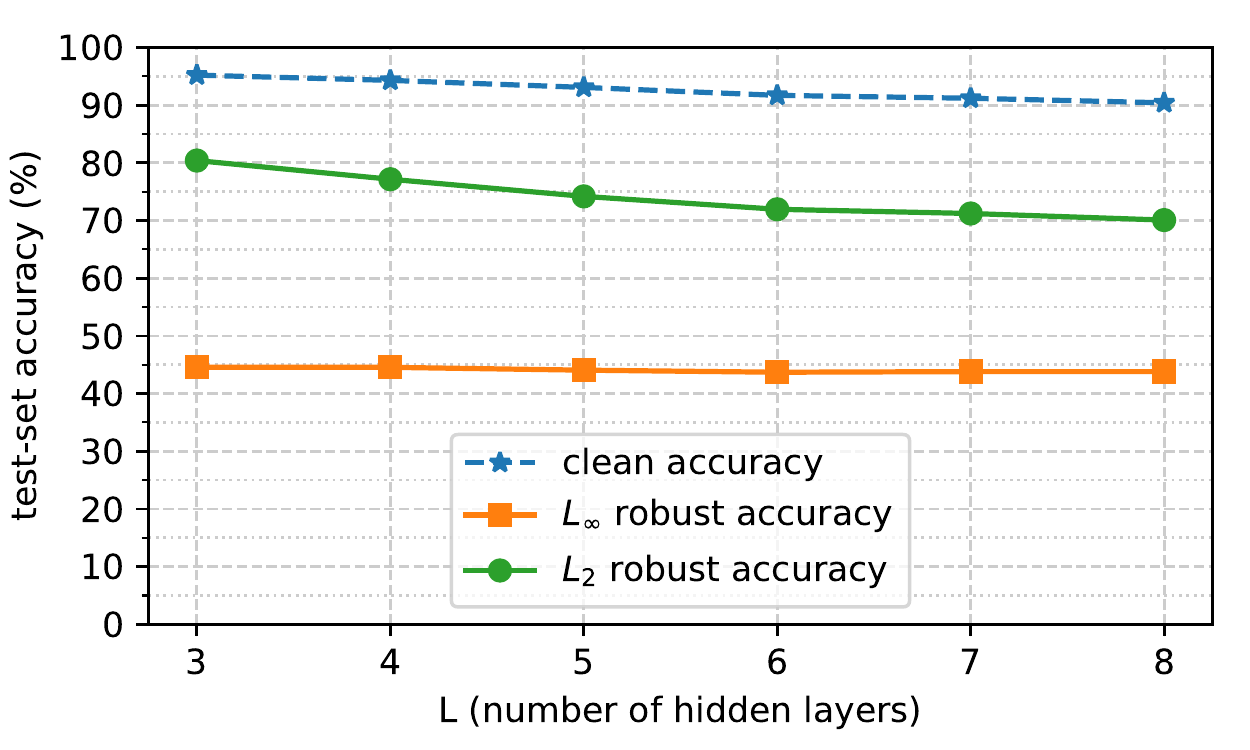}
\end{minipage}\qquad\begin{minipage}[h]{0.45\linewidth}
\centering
\resizebox{1\linewidth}{!}{%
\centering
\begin{tabular}{|l|l|l|l|l|l|l|l|}
\hline
\multicolumn{2}{|L{1.6cm}|}{}  & {\bf $L=3$}   & {\bf $L=4$}   & {\bf $L=5$}   & {\bf $L=6$}   & {\bf $L=7$}   & {\bf $L=8$}\\
\hline
\multicolumn{2}{|L{1.6cm}|}{\it Clean Accuracy:}    &   95.21   &   94.30   &   93.08   &   91.70   &   91.20   &   90.41  \\
\hline\hline
$L_\infty$  &   \pgd                                &   44.57   &   44.55   &   44.06   &   43.70   &   43.81   &   43.81    \\
\hline\hline
        $L_2$       &   \pgd                        &	80.40	&	77.16	&	74.19	&	71.95	&	71.22	&	70.08	\\
\hline
\end{tabular}%
}
\end{minipage}
\caption{CIFAR-$10$ test-set clean and robust accuracies (under transfer attacks) of LINAC defended classifiers with different numbers of implicit network layers $L$, keeping all other hyper-parameters fixed.\label{figs:linac_sensitivity_L}}
\end{figure}

In Figure~\ref{figs:linac_sensitivity_L}  we vary the number of implicit network layers $L$, keeping all other hyper-parameters the same, including the representation layer index $K=2$ and number of epochs $N=10$. Note that we used $L=5$ for our defended classifier evaluated in the main paper.

\begin{figure}[h]
\centering
\begin{minipage}[h]{0.45\linewidth}
\centering
\includegraphics[width=0.95\linewidth]{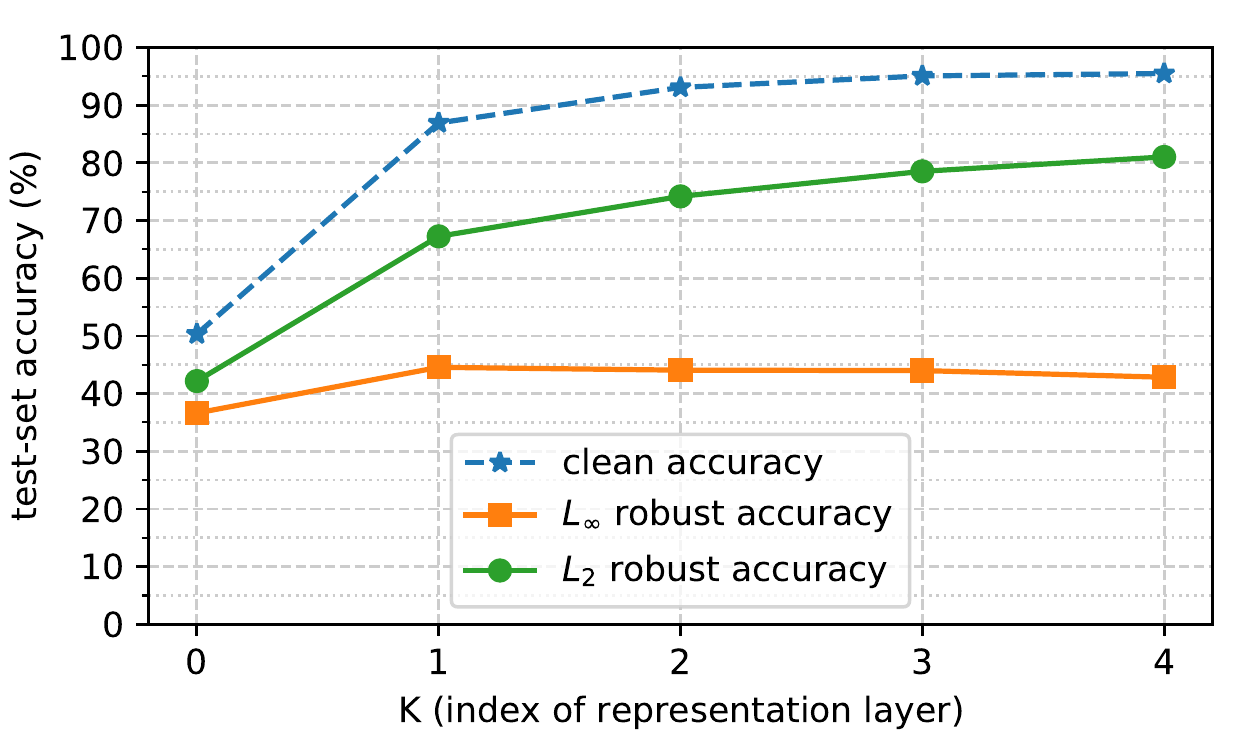}
\end{minipage}\qquad\begin{minipage}[h]{0.45\linewidth}
\centering

\resizebox{1\linewidth}{!}{%
\centering
\begin{tabular}{|l|l|l|l|l|l|l|}
\hline
\multicolumn{2}{|L{1.6cm}|}{}  &	$K=0$	&	$K=1$	&	$K=2$	&	$K=3$	&	$K=4$	\\
\hline
\multicolumn{2}{|L{1.6cm}|}{\it Clean Accuracy:}    &	50.32	&	86.92	&	93.08	&	95.07	&	95.49	\\
\hline\hline
$L_\infty$  &   \pgd                                &	36.63	&	44.55	&	44.06	&	43.99	&	42.83	\\
\hline\hline
        $L_2$       &   \pgd                        &	42.17	&	67.26	&	74.19	&	78.53	&	81.03	\\
\hline
\end{tabular}%
}
\end{minipage}
\caption{CIFAR-$10$ test-set clean and robust accuracies (under transfer attacks) of LINAC defended classifiers with different implicit network layers used to output representations ($K$), keeping all other hyper-parameters the same.\label{figs:linac_sensitivity_K}}
\end{figure}

In Figure~\ref{figs:linac_sensitivity_K} we change the index of the LINAC representation layer $K$, keeping all other hyper-parameters unchanged. Note that we used $K=2$ for our defended classifier evaluated in the main paper.

\begin{figure}[h]
\centering
\begin{minipage}[h]{0.45\linewidth}
\centering
\includegraphics[width=0.95\linewidth]{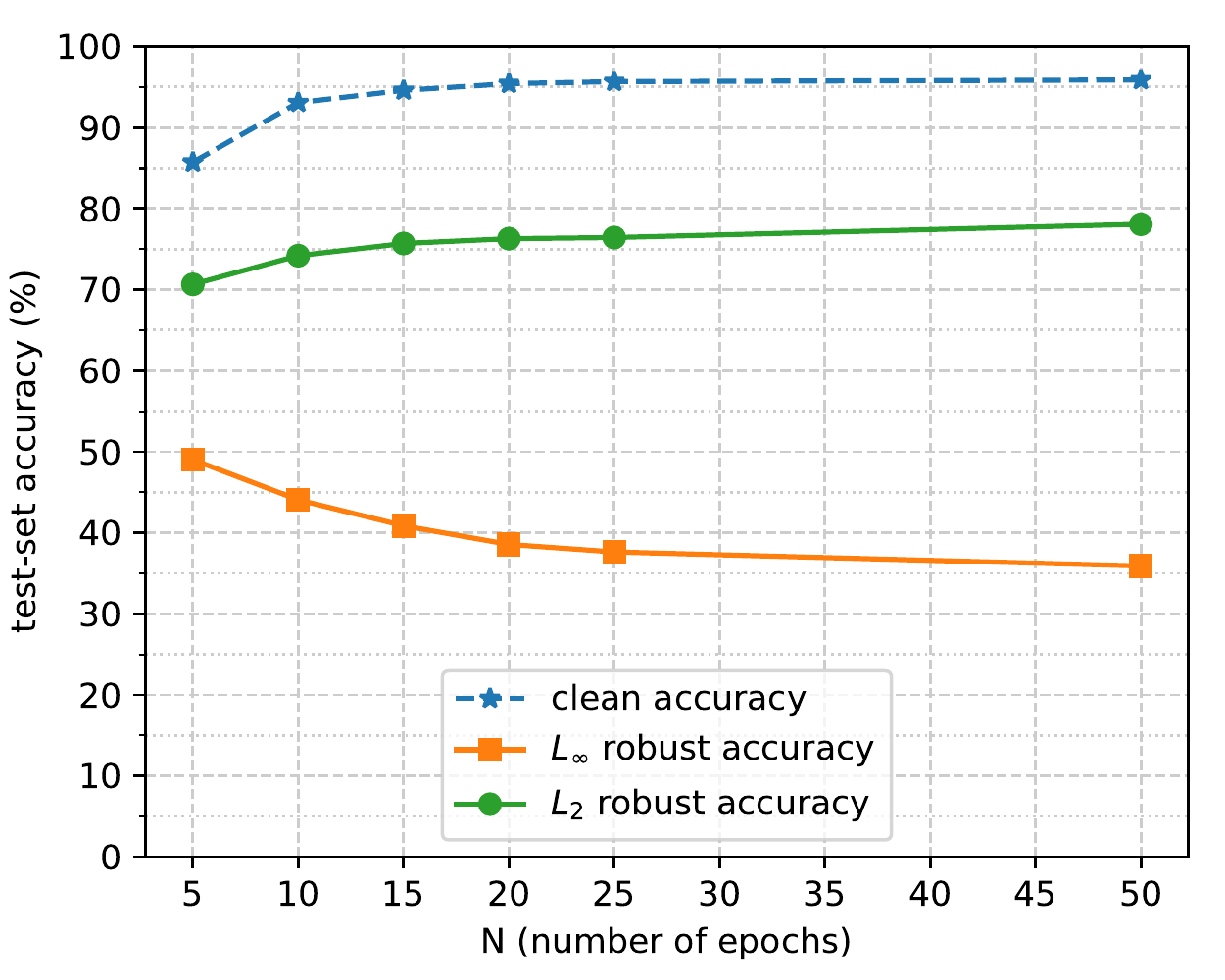}
\end{minipage}\qquad\begin{minipage}[h]{0.45\linewidth}
\centering
\resizebox{1\linewidth}{!}{%
\centering
\begin{tabular}{|l|l|l|l|l|l|l|l|}
\hline
\multicolumn{2}{|L{1.6cm}|}{}  &	$N=5$	&	$N=10$	&	$N=15$	&	$N=20$	&	$N=25$	&	$N=50$\\
\hline
\multicolumn{2}{|L{1.6cm}|}{\it Clean Accuracy:}    &	85.74	&	93.08	&	94.59	&	95.41	&	95.66	&	95.87	\\
\hline\hline
$L_\infty$  &   \pgd                                &	49.01	&	44.06	&	40.86	&	38.56	&	37.63	&	35.90	\\
\hline\hline
        $L_2$       &   \pgd                        &	70.65	&	74.19	&	75.69	&	76.28	&	76.42	&	78.06	\\
\hline
\end{tabular}%
}
\end{minipage}
\caption{CIFAR-$10$ test-set clean and robust accuracies (under transfer attacks) of LINAC defended classifiers with implicit networks trained for $N$ epochs, keeping all other hyper-parameters constant.\label{figs:linac_sensitivity_N}}
\end{figure}

In Figure~\ref{figs:linac_sensitivity_N} we analyse the sensitivity of LINAC to the number of epochs $N$, keeping all other hyper-parameters constant. Note that we used $N=10$ for our defended classifier evaluated in the main paper.

\section{Characterising the LINAC Transform}
\label{sec:lossy_encoding}

\begin{figure}[h]
\centering
\includegraphics[width=0.6\linewidth]{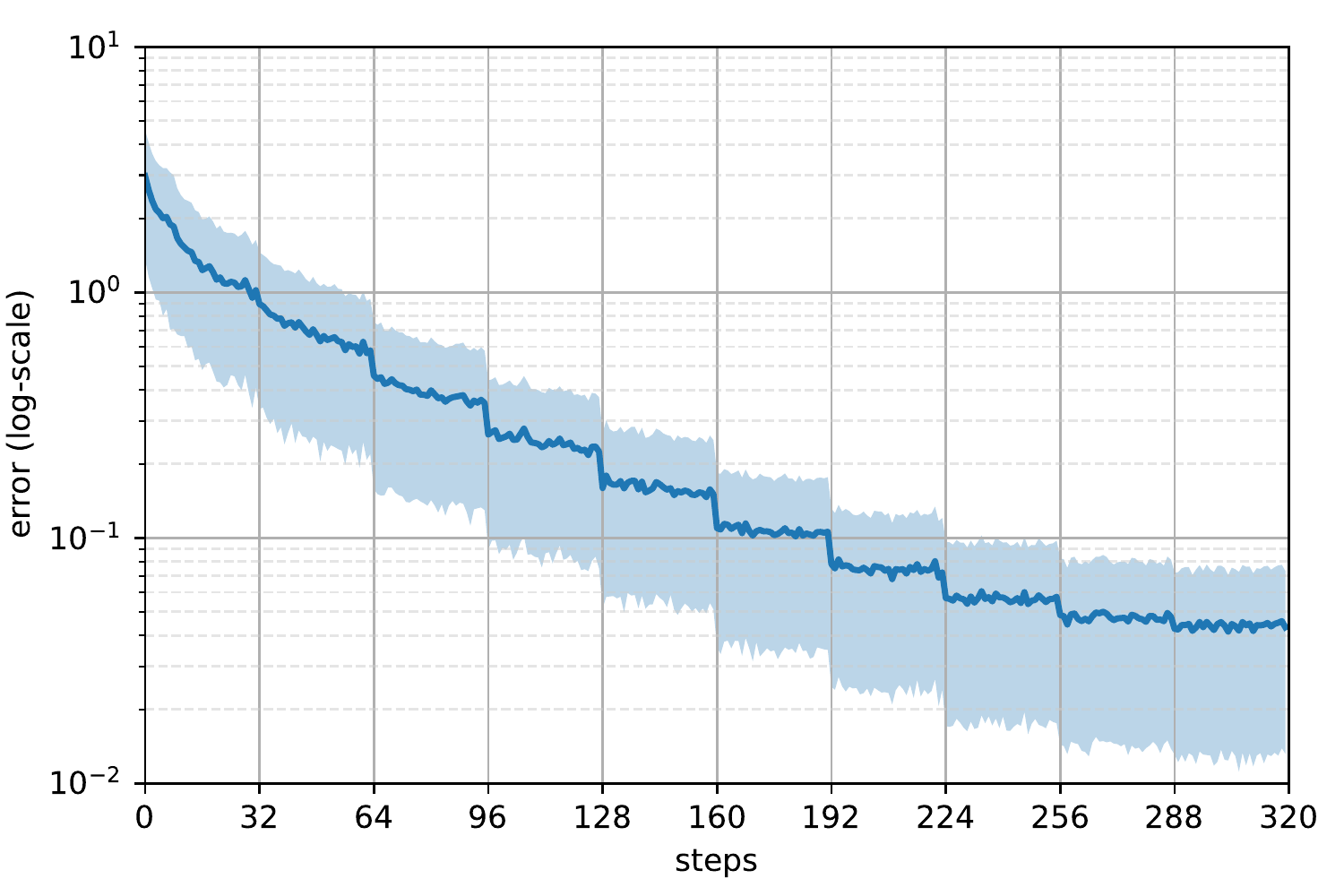}
\caption{Independent fitting of implicit neural networks to CIFAR-$10$ test-set images in order to compute their LINAC transforms. Sum squared encoding errors, averaged over pixels, are plotted against fitting steps. \label{figs:linac_learning_curves}}
\end{figure}

In Figure~\ref{figs:linac_learning_curves} we plot learning curves characterising implicit network fitting, as used for our defended classifier. Mean and standard deviation of errors across independent learning processes for the entire CIFAR-$10$ test-set are plotted as functions of optimisation steps, using a log-scale for errors. The final mean value of such errors is $0.04325$, which confirms that our LINAC approach leads to lossy representations.

\begin{figure}[h]
\centering
\includegraphics[width=0.5\linewidth]{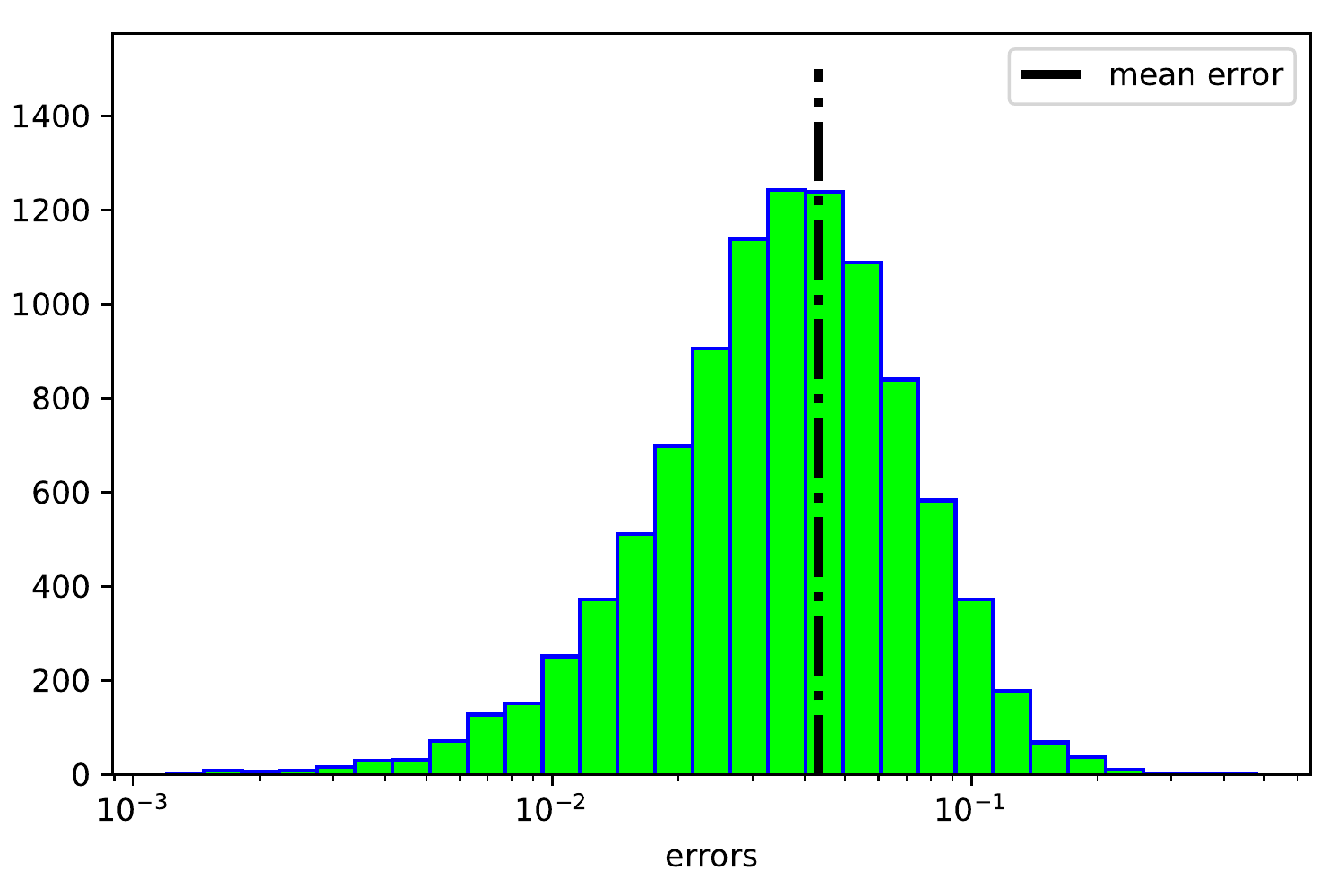}
\caption{Histogram of LINAC transform encoding errors plotted for the entire CIFAR-$10$ test-set. The overall mean value of such errors is $0.04325$, which confirms that our LINAC approach leads to lossy representations.\label{figs:linac_error_hist}}
\end{figure}

A histogram of final sum squared errors for the entire test-set of CIFAR-$10$ is provided in Figure~\ref{figs:linac_error_hist}.

\begin{figure}[h]
\centering
\includegraphics[width=0.35\linewidth]{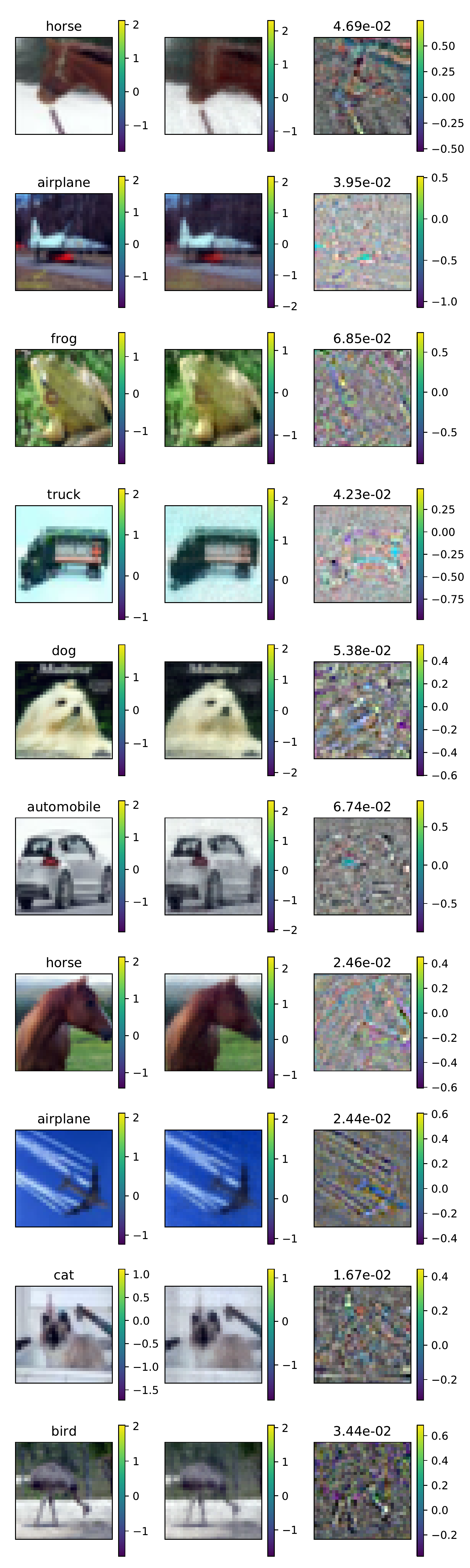}

\caption{Image approximations computed for LINAC with the \privatekey, as used for our defended classifier. Original images and labels are plotted in the first column. Note that labels are not used for LINAC. Implicit network outputs are plotted in the second column. Difference images and sum squared errors, averaged over pixels, are plotted in the third column. Note that LINAC uses lossy image approximations.\label{figs:linac_example_privatekey}}
\end{figure}

\begin{figure}[h]
\centering

\includegraphics[width=0.35\linewidth]{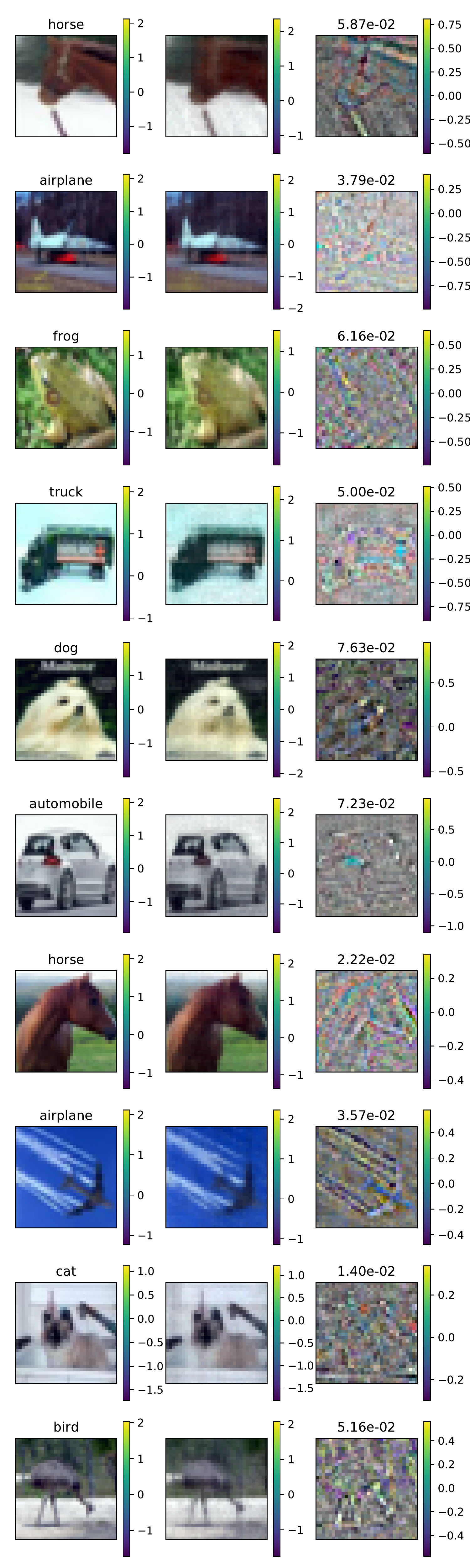}

\caption{Image approximations computed for LINAC with a \textbf{different}, attacker chosen key. Original images and labels are plotted in the first column. Note that labels are not used for LINAC. Implicit network outputs are plotted in the second column. Difference images and sum squared errors, averaged over pixels, are plotted in the third column. Note that LINAC leads to lossy image approximations which are key dependent.\label{figs:linac_example_otherkey}}
\end{figure}

For a qualitative evaluation of such statistics, we provide examples of original images, their reconstructions and difference images, using LINAC and the \privatekey in Figure~\ref{figs:linac_example_privatekey} and, for comparison, a different key in Figure~\ref{figs:linac_example_otherkey}. We observe that encoding errors using LINAC are key dependent. Furthermore, significant amounts of  information seem to be left out by LINAC. Some difference images could be recognised as the correct class, most likely due to high-frequency information which is not well represented.

\begin{figure}[h]
\centering
\includegraphics[width=0.5\linewidth]{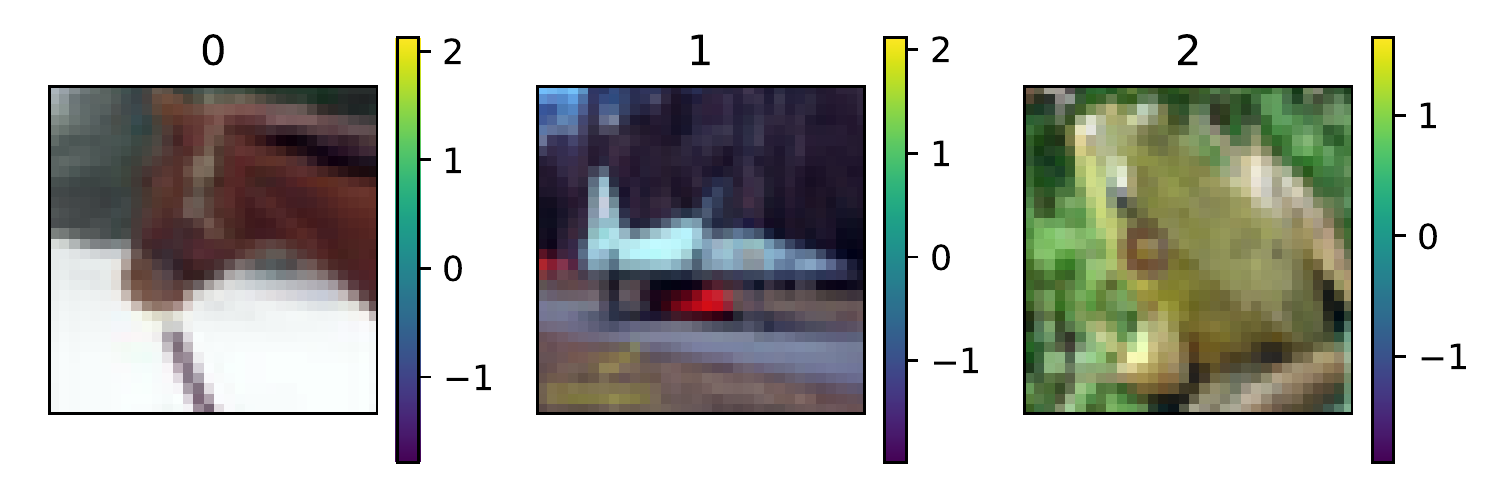}

\includegraphics[width=\linewidth]{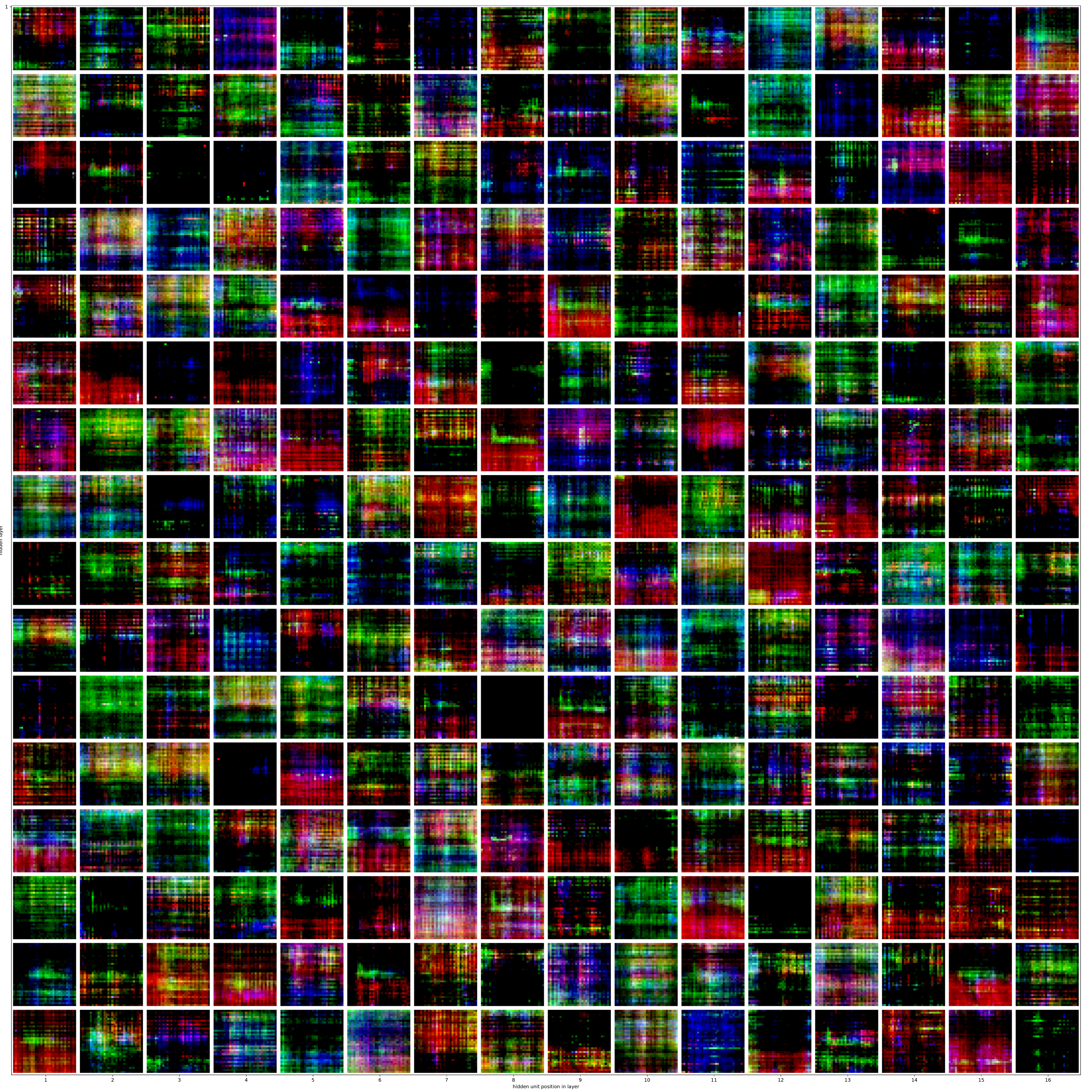}

\caption{Comparing transforms of the 3 top images using LINAC with the \privatekey, as done for our defended classifier. The respective activation images with $H=256$ channels were plotted in a $16 \times 16$ grid of slices of the same size with original images. Respective slices over the channel dimension of activation images were combined as RGB channels in this plot (bottom), in order to compare channel representations for the three input images (top). Each square in the grid represents the activations of a LINAC representation channel for all pixels in the original image. Different values of RGB signify differences in LINAC representations across images.\label{figs:linac_3examples_privatekey}}
\end{figure}

\begin{figure}[h]
\centering
\includegraphics[width=0.5\linewidth]{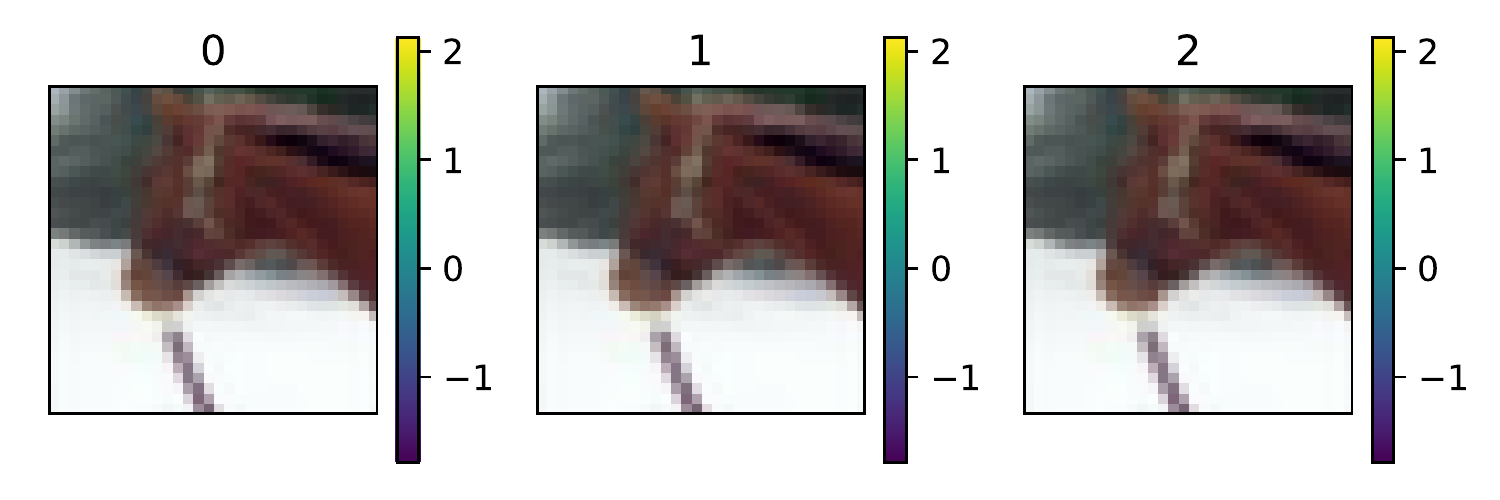}

\includegraphics[width=\linewidth]{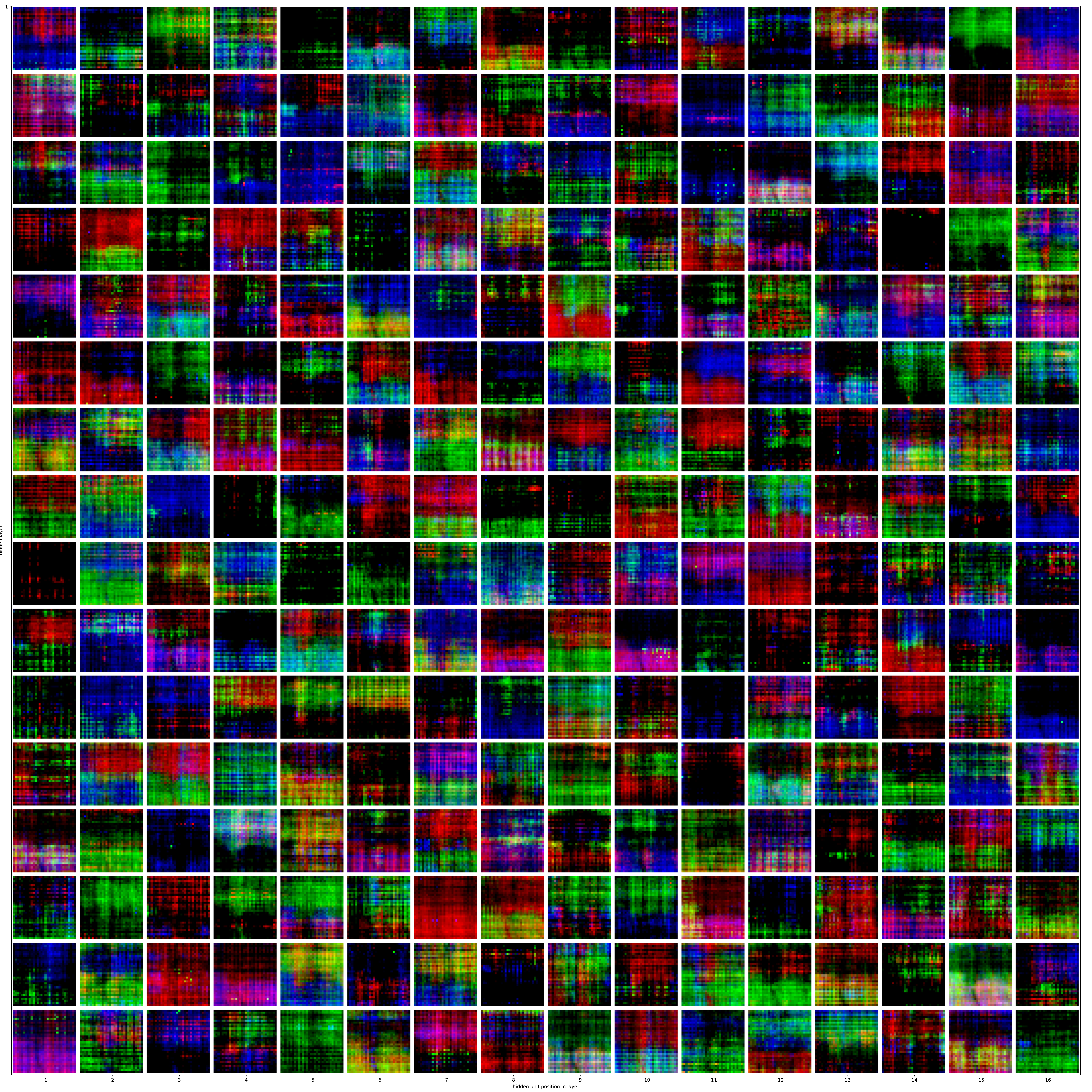}

\caption{Comparing LINAC transforms of the same image using the \privatekey and two other random keys. The respective activation images with $H=256$ channels were plotted in a $16 \times 16$ grid of slices of the same size with original images. Respective slices over the channel dimension of resulting activation images were combined as RGB channels in this plot (bottom), in order to compare channel representations with three different keys for the same input image. Each square in the grid represents the activations of a LINAC representation channel for all pixels in the original image. Different values of RGB signify differences in LINAC representations across keys. \label{figs:linac_example0_keys}}
\end{figure}

\begin{figure}[h]
\centering
\includegraphics[width=0.5\linewidth]{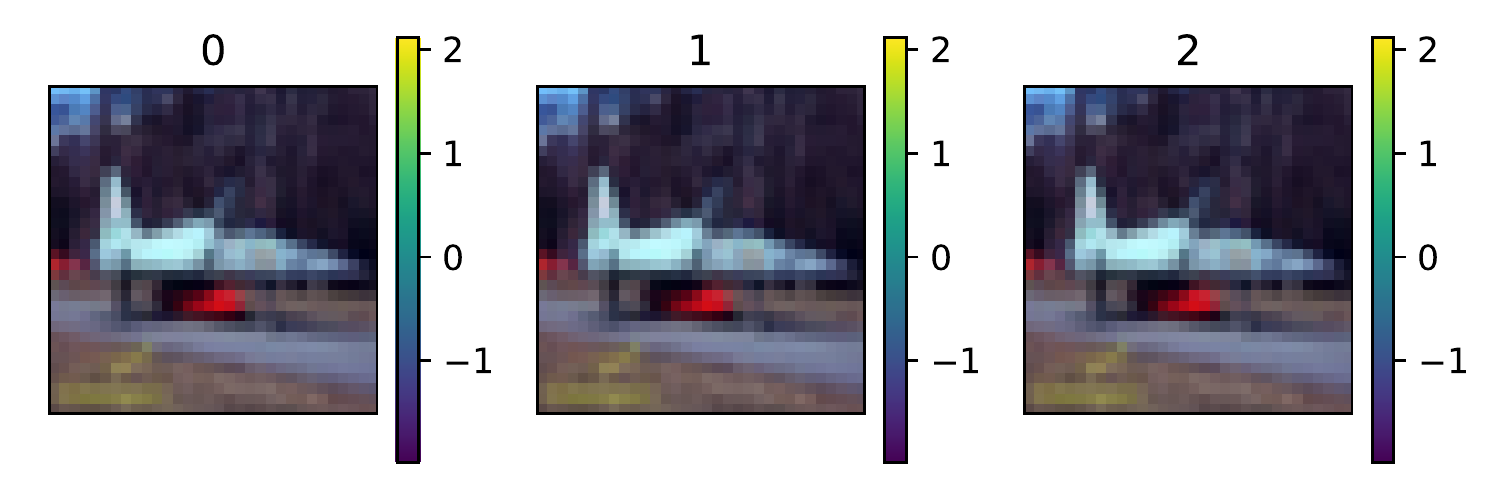}

\includegraphics[width=\linewidth]{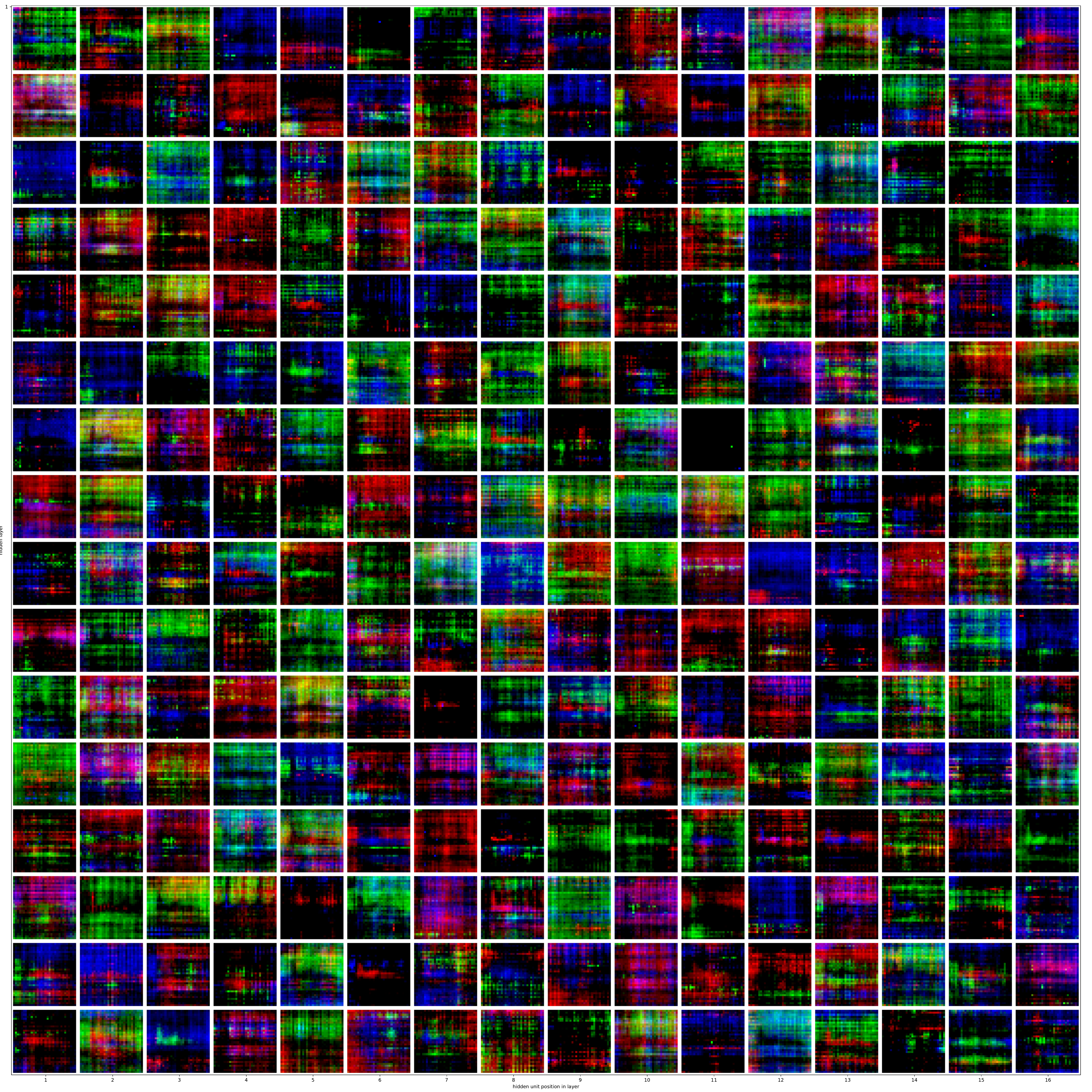}

\caption{Comparing LINAC transforms of the same image using the \privatekey and two other random keys. The respective activation images with $H=256$ channels were plotted in a $16 \times 16$ grid of slices of the same size with original images. Respective slices over the channel dimension of resulting activation images were combined as RGB channels in this plot (bottom), in order to compare channel representations with three different keys for the same input image. Each square in the grid represents the activations of a LINAC representation channel for all pixels in the original image. Different values of RGB signify differences in LINAC representations across keys. \label{figs:linac_example1_keys}}
\end{figure}

\begin{figure}[h]
\centering
\includegraphics[width=0.5\linewidth]{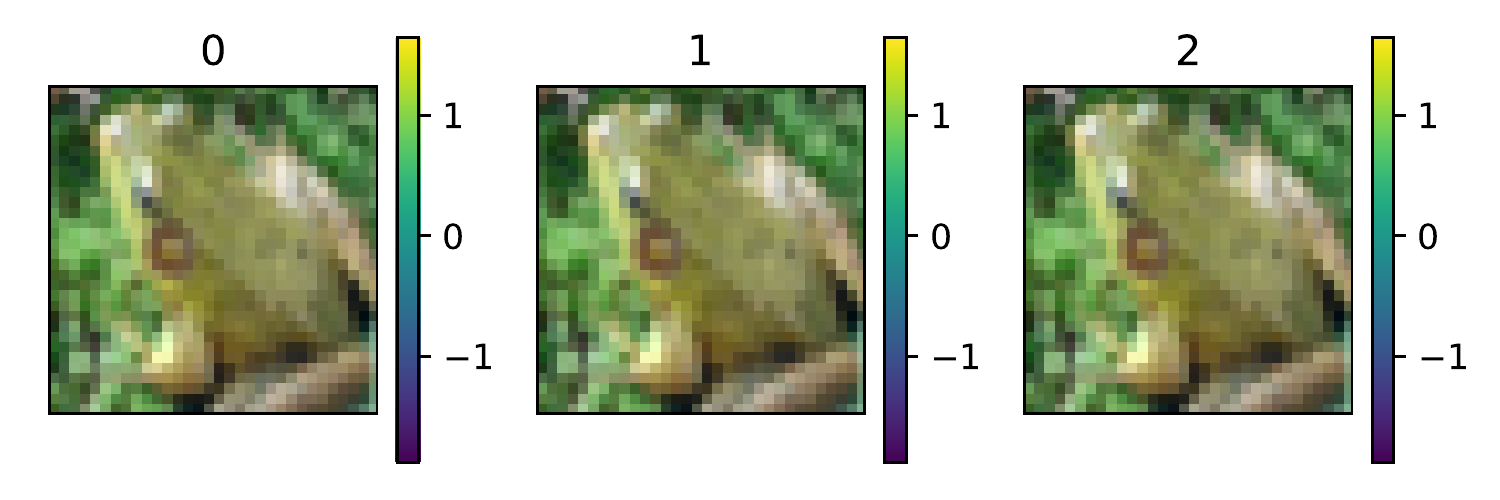}

\includegraphics[width=\linewidth]{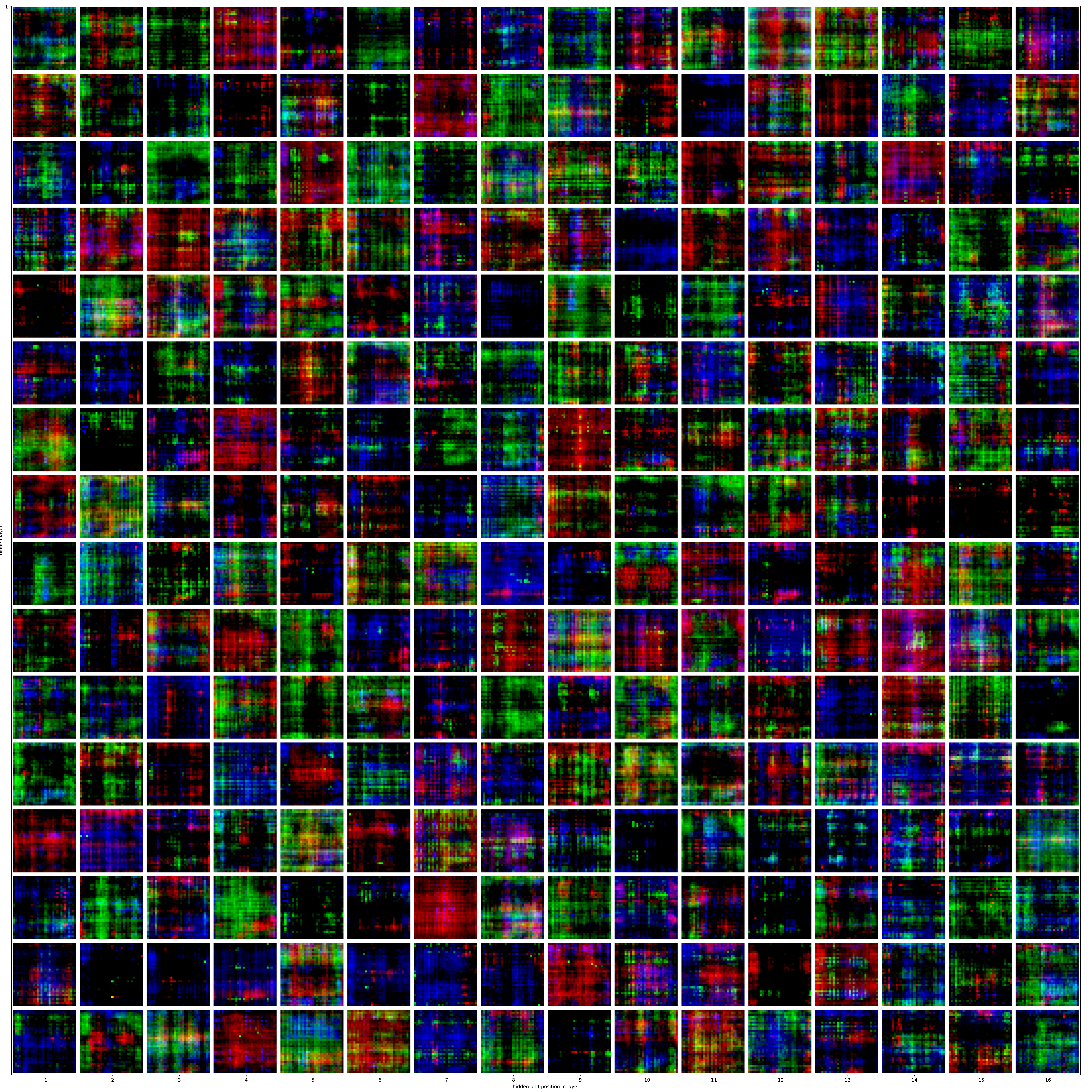}

\caption{Comparing LINAC transforms of the same image using the \privatekey and two other random keys. The respective activation images with $H=256$ channels were plotted in a $16 \times 16$ grid of slices of the same size with original images. Respective slices over the channel dimension of resulting activation images were combined as RGB channels in this plot (bottom), in order to compare channel representations with three different keys for the same input image. Each square in the grid represents the activations of a LINAC representation channel for all pixels in the original image. Different values of RGB signify differences in LINAC representations across keys. \label{figs:linac_example2_keys}}
\end{figure}

Finally, we provide a number of plots for qualitative comparisons of LINAC transforms. Figure~\ref{figs:linac_3examples_privatekey} shows three different images encoded and their respective LINAC representations encoded as RGB channels. Figures \ref{figs:linac_example0_keys}, \ref{figs:linac_example1_keys} and \ref{figs:linac_example2_keys} plot LINAC transforms of the same respective images, but using with different keys, one on each RGB colour channel.

\newpage

\newpage

\newpage

\newpage

\newpage

\newpage

\end{document}